\pdfoutput=1

\documentclass[11pt]{article}

\usepackage[]{EMNLP2022}

\usepackage{times}
\usepackage{latexsym}
\usepackage{amsmath}
\usepackage{amssymb}
\usepackage{bm}
\usepackage{graphicx}
\usepackage{float}
\usepackage{booktabs}     
\usepackage{subcaption}%

\usepackage[T1]{fontenc}

\usepackage[utf8]{inputenc}

\usepackage{microtype}

\usepackage{inconsolata}

\title{\textit{Scaling Laws vs Model Architectures}: \\ How does Inductive Bias Influence Scaling?}

\author{Yi Tay$\thanks{\: Yi and Mostafa contributed equally. Samira is now at Apple.}$ \hspace{5mm} Mostafa Dehghani$^*$ \hspace{5mm}  Samira Abnar \hspace{5mm}  Hyung Won Chung \hspace{5mm} \\
\textbf{William Fedus \hspace{5mm} Jinfeng Rao \hspace{5mm}  Sharan Narang \hspace{5mm} Vinh Q. Tran }\\  \vspace{2mm} 
\textbf{Dani Yogatama$^\dagger$ \hspace{5mm} Donald Metzler} \\ \vspace{2mm}
  Google Research \& DeepMind$^\dagger$ \\
  \texttt{\{yitay,dehghani\}@google.com} \\
  }
\begin{document}
\maketitle
\begin{abstract}
There have been a lot of interest in the scaling properties of Transformer models \citep{kaplan2020scaling}. However, not much has been done on the front of investigating the effect of scaling properties of different inductive biases and model architectures. Do model architectures scale differently? If so, how does inductive bias affect scaling behaviour?  How does this influence upstream (pretraining) and downstream (transfer)? This paper conducts a systematic study of scaling behaviour of ten diverse model architectures such as Transformers, Switch Transformers, Universal Transformers, Dynamic convolutions, Performers, and recently proposed MLP-Mixers. Via extensive experiments, we show that (1) architecture is an indeed an important consideration when performing scaling and (2) the best performing model can fluctuate at different scales. We believe that the findings outlined in this work has significant implications to how model architectures are currently evaluated in the community.
\end{abstract}

\section{Introduction}
There have been a lot recent interest in the scaling properties of Transformer models \citep{kaplan2020scaling,hernandez2021scaling,bahri2021explaining,henighan2020scaling,tay2021scale, abnar2021exploring}. However, not much is understood about the scaling properties of different inductive biases imposed by model architectures. Improvements at a a specific scale (compute, size etc) are often assumed to transfer to different scales and compute regions \citep{so2019evolved,choromanski2020rethinking,lan2019albert,dehghani2018universal} and new research is often presented in a point-wise fashion with respect to scale. In short, it is not uncommon for new methods to be presented with data points at very specific or limited compute regions (e.g., base size). We believe that understanding the interaction between architecture and scaling laws is crucial as designing models that perform well at diverse scales will likely have significant impact.

This paper is an attempt to understand the effect of inductive bias (architecture) on scaling laws of language models. To this end, we pre-train and finetune over ten diverse model architectures across multiple compute region and scales (e.g., from 15M to 40 Billion parameters). In total, we pre-train and finetune over 100 different models of different architectures and sizes and present insights and challenges at scaling these ten diverse architectures. 

\begin{figure*}[]
     \centering
     \begin{subfigure}[]{0.48\textwidth}
         \centering
         \includegraphics[width=1.0\textwidth]{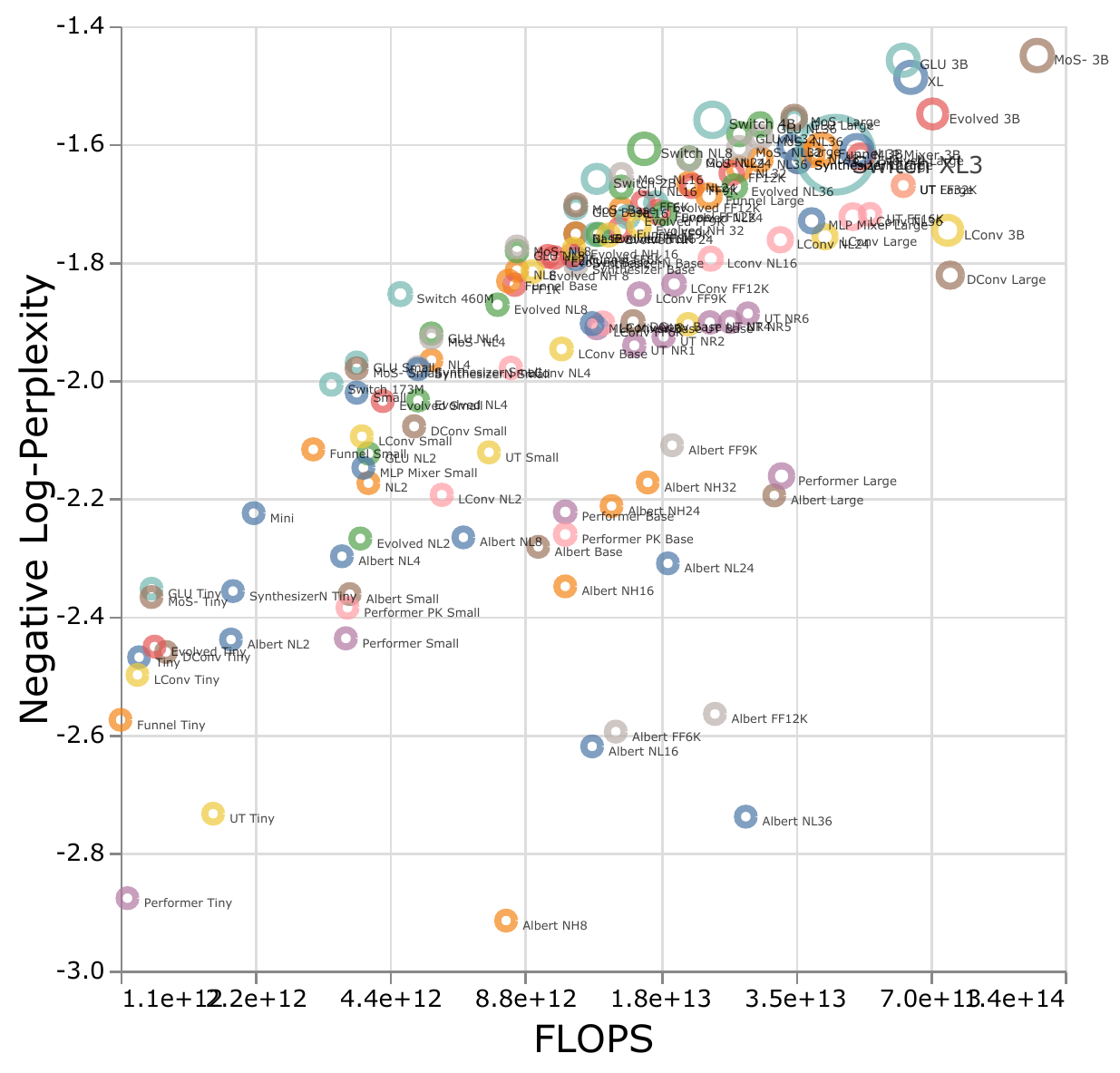}
         \caption{Upstream: Negative Log-Perplexity}
         \label{fig:overview_us}
     \end{subfigure}
     \begin{subfigure}[]{0.48\textwidth}
         \centering
         \includegraphics[width=1.0\textwidth]{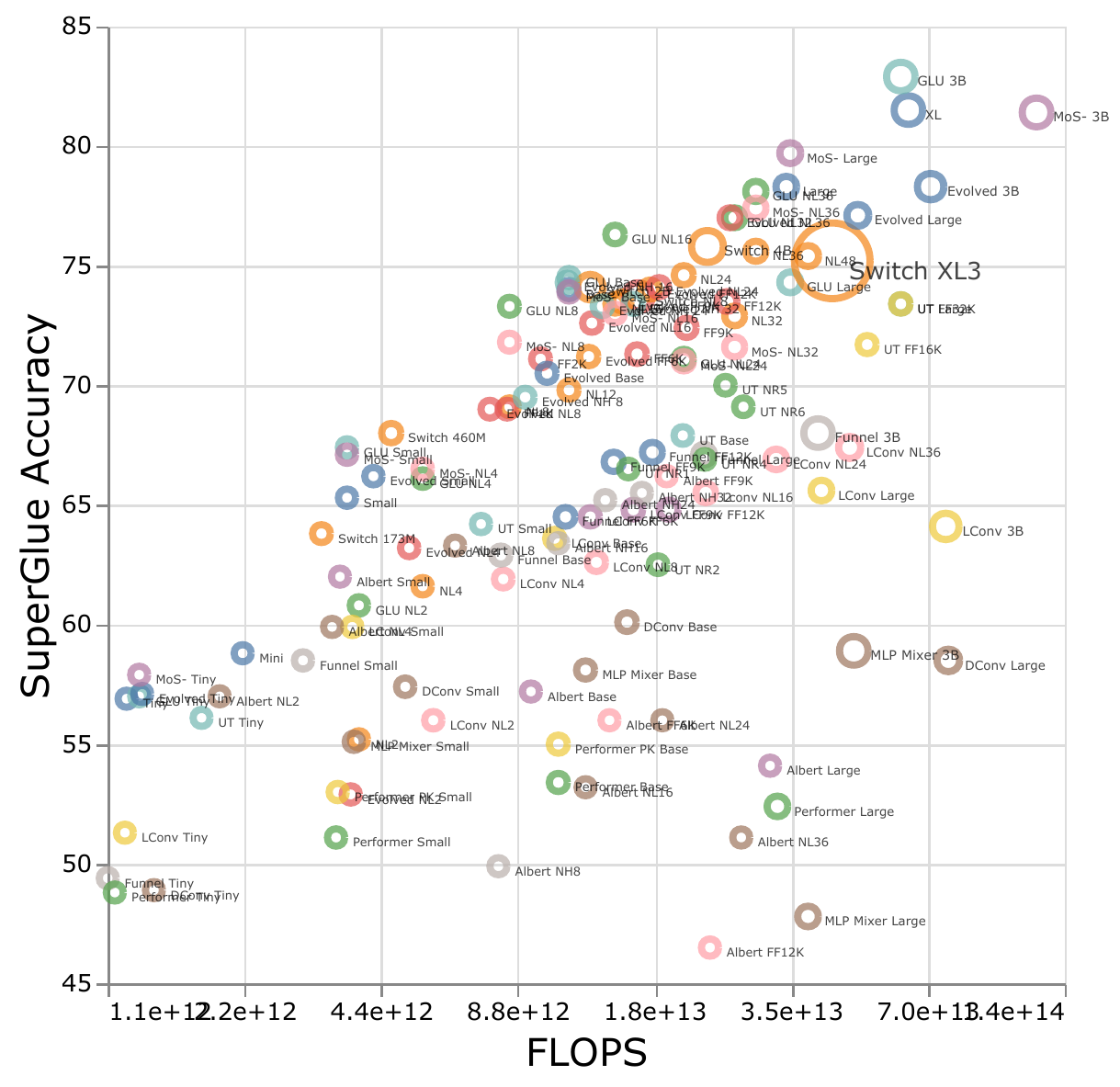}
         \caption{Downstream: Accuracy}
         \label{fig:overview_ds}
     \end{subfigure}
    \caption{An overview compute-performance (FLOPs vs performance) plot of all the diverse models and architectures we pretrained and finetuned in this study. Colors represent different model architectures and size of the circles represent the size of the model (parameters).}
    \label{fig:overview}
\end{figure*}

We consider a broad spectrum of models in our extensive experiments. Concretely, we consider several well-established Transformer variants \citep{vaswani2017attention} such as Evolved Transformer \citep{so2019evolved}, Universal Transformers \citep{dehghani2018universal} and Switch Transformers \citep{fedus2021switch}. We also consider lightweight models such as ALBERT \citep{lan2019albert} and/or efficient Transformers \citep{tay2020efficient} such as Performer \citep{choromanski2020rethinking} and Funnel Transformers \citep{dai2020funnel}. In our comparison, we are also interested in finding out if general improvements to the Transformer architectures such as Mixture-of-Softmax \citep{yang2017breaking} and/or Gated Linear Units \citep{dauphin17_glu,shazeer2020glu} influence the scaling behaviour of models. Finally, we also evaluate models outside the family of Transformers including Lightweight convolutions \citep{wu2019pay}, Dynamic convolutions \citep{wu2019pay} and the recently proposed MLP-Mixers \citep{tolstikhin2021mlp}. Figure \ref{fig:overview} illustrates an overview about the experiments we run. 

We also note that scaling these models is not as straightforward as it seems, i.e., there are intricate details of scale that are intertwined with architectural choices which we study in detail in this paper. For example, a distinct feature of Universal Transformers (and ALBERT) is parameter sharing. Hence, compared with standard Transformers, this architectural choice significantly warps the scaling behaviour not only with respect to performance but also amongst compute metrics such as FLOPs, speed and number of parameters~\citep{dehghani2021efficiency}. Conversely, models such as Switch Transformers are on the other end of the spectrum with an uncommon relationship between FLOPs and number of parameters, i.e., they have high parameter to FLOPs ratio. This difficulty makes navigating this landscape challenging. 

\paragraph{Our Contributions and Insights} The key contributions of this paper are as follows:
\begin{itemize}
    \item For the first time, we derive scaling laws for different inductive biases and model architectures. We find that this scaling coefficient differs greatly from model to model. We believe this is an important consideration in model development. It turns out that amongst all ten architectures that we consider, the vanilla Transformer has the best scaling behaviour, even if its absolute performance at each compute region is not the greatest. 
    \item We observe that models that operate well in one compute-scale region is not necessarily the best in another compute-region. Moreover, we find that certain models have difficulty scaling despite performing decently (comparably) at lower-compute regions. This has implications, since it is difficult to get the fulll picture of a model's scalability with pointwise comparisons at a certain compute-region.
    \item We find that when it comes to scaling different model architectures, upstream pre-training perplexity might not correlate well with downstream transfer. Hence, the underlying architecture and inductive bias is also crucial for downstream transfer.
    \item We highlight the difficulties of scaling with certain architectures and show that some models do not scale (or scale with a negative trend). We also find concerning trends where linear-time attention models such as Performer struggle with scaling up. 
\end{itemize}

\section{Related Work}
\citet{kaplan2020scaling} studied empirical scaling laws of the decoder-only Transformer language models. They focused on the standard left-to-right language modeling objective with the cross-entropy loss as the performance metric. One of the main findings is that the loss scales as a power-law with three major characteristics of the model training: model size, dataset size and the training compute. Another somewhat surprising finding is that the model shapes such as width or depth of the Transformer network have minimal effects on the cross-entropy loss for a wide range of scales. Subsequent works~\citep{henighan2020scaling,hernandez2021scaling} made similar conclusions for autoregressive generative modeling and for transfer learning, respectively. This finding is also generally supported by \citep{tay2021scale} but discrepancies were found for the gap between pretraining and finetuning - highlighting the fact that observing downstream performance of large language model is indeed important. In \citep{tay2021scale}, the effect of depth was unusually pronounced for downstream performance.

\citet{raffel2019exploring} studied the effect of pre-training objectives, model structures (e.g., encoder-decoder, decoder-only), pre-training dataset size and training strategy on the transfer learning. They showed that the downstream performance monotonically increases with the model scale (from 60M to 11B parameters). While they studied several model structures, the Transformer implementation is mostly the same as the original Transformer by~\citet{vaswani2017attention}. \citet{conneau2020unsupervised,goyal2021} scaled-up multilingual encoder-only architectures up to 11B parameters while maintaining the original Transformer implementation. They found that scaling the model improves its cross-lingual ability. \citet{fedus2021switch} scaled a sparse model based on Mixture of Experts (MoE) models up to trillion parameters.

While previous studies have repeatedly shown the benefits of scale for language understanding tasks for both dense and sparse Transformers and cross-lingual abilities, all of these used the same Transformer implementation within each studies. With a plethora of improved Transformer architectures proposed in the literature, it is timely to investigate which of these improved architecture has the best scaling properties. The main goal of this paper is to systematically study how inductive biases imposed by these Transformer variants affect the scaling behavior in a shared software and hardware settings. This is in similar spirit to \citep{narang2021transformer} that studies the impact of architectures on performance. Our analysis extends that of \citep{narang2021transformer} to the model scale axis. 


\section{Methods}
This section outlines our experimental setup. 
\subsection{Models}
This section describes the models we evaluate in our experiments. Our models are largely implemented in a sequence to sequence framework \citep{sutskever2014sequence} following the convention of T5 \citep{raffel2019exploring}. Encoder-decoder models are a natural choice for this experimentation because they can universally express both encoding and decoding tasks.  
\paragraph{Transformer Variants} We consider several standard Transformer variants.

\begin{itemize}
    \item \textbf{Transformers} \citep{vaswani2017attention} - The basic vanilla Transformer architecture. Our basic setup considers the T5-style of Transformers \citep{raffel2019exploring}, which largely follows the vanilla Transformer except that it uses relative attention instead of sinusoidal position embeddings and pre-layer normalization, i.e. layer normalization is applied \textit{before} each sublayer.
    \item \textbf{Evolved Transformers} \citep{so2019evolved} - A transformer architecture learned via AutoML. The architecture comprises of convolutions and attention. We scale Evolved Transformers following the same pattern as vanilla Transformers. 
    \item \textbf{Universal Transformers (UT)} \citep{dehghani2018universal} - A Transformer architecture with shared parameters and recurrent-like computation for transform layers. Scaling UTs are challenging because of parameter sharing. While we are able to also increase $d_{FF}$ or $d_{model}$, the increase in parameters is of magnitude $N_{layers}$ than standard Transformers. Another axis of exploration is to scale $r$ the number of repeated computation at each UT layer - this increases computation (number of FLOPs) but does not increase the parameter size of the model.
    \item \textbf{Switch Transformer} \citep{fedus2021switch} - a sparsely activated mixture-of-experts architecture. The Sparse Transformer is another model with an unusual relationship between number of parameters and compute. When we scale this model uniformly, the number of parameters easily reaches the ballpark of 40B.
 \end{itemize}
\paragraph{Efficient Transformer Variants} These class of models are mainly concerned at reducing computational costs, memory usage, or parameter count of models.
\begin{itemize}
    \item \textbf{Performer} \citep{choromanski2020rethinking} - A linear time attention model using generalizable kernel attention. For simplicity, we adopt the relu kernel variant for our experiments.  We scale Performer in the similar fashion (i.e., uniform scaling) as vanilla Transformers.
    \item \textbf{Funnel Transformer (FT)} \citep{dai2020funnel} A Transformer architecture that downsamples the input sequence across the layer stack. Our implementation uses FT only in the encoder and reverts to vanilla Transformer in the decoder following \citet{narang2021transformer}.
       \item \textbf{ALBERT} \citep{lan2019albert} - A lightweight transformer architecture that shares parameters across all layers and factorizes the embedding and output softmax layers. For our seq2seq ALBERT, we also share the weights of encoder and decoder.
\end{itemize}
\paragraph{General Improvements} We consider general improvements that are not necessarily tied to Transformers. We select candidates that have shown to do well in \citet{narang2021transformer}.
\begin{itemize}
    \item \textbf{Mixture of Softmaxes} \citep{yang2017breaking} - A transformer architecture adopting the MoS method at the Softmax layer.
    \item \textbf{Gated Linear Units with GeLU} (GLU-Transformer) - Replacing position-wise feed-forward-networks in Transformers with Gated Linear Units~\citep{dauphin17_glu}.
\end{itemize}
\paragraph{Non-Transformer Architectures} We are interested in the scaling behaviour of non-Transformer based architectures such as convolutions and/or mixer architectures. 
\begin{itemize}
    \item \textbf{Lightweight Convolutions} \citep{wu2019pay} - Lightweight depthwise convolutions that have shown promise over Transformer architectures. 
    \item \textbf{Dynamic Convolutions} \citep{wu2019pay} - An extension of the Lightweight Convolution to create time-dependent kernels. 
    \item \textbf{MLP-Mixers} \citep{tolstikhin2021mlp} - Mixers are recently proposed architectures that learn a lightweight mixing of tokens. Since Mixers have not been used in autoregressive decoding, we only use token-mixers on the input encoder.
 \end{itemize}

 \subsection{Experiment Setup}
 Our setup, along with all models, are implemented in Mesh TensorFlow~\citep{shazeer2018mesh}, a library with similar interface to TensorFlow but enables distributed model parallelism across multiple workers. For fair comparison, all models are pretrained for $2^{19}$ steps on the english C4 corpus optimized using an inverse square root learning rate with Adafactor \citep{shazeer2018adafactor}. All models use the same SentencePiece tokenizer \citep{kudo2018sentencepiece} containing $32K$ subwords. This closely follows the setup in the T5 paper \citep{raffel2019exploring}. Finetuning is performed for $100K$ steps on a mixture of GLUE \citep{wang2018glue}, SuperGLUE \citep{wang2019superglue} and SQuAD \citep{rajpurkar2016squad}. We evaluate on both upstream (pre-training) validation perplexity as well as downstream transfer for NLU tasks (GLUE + SuperGLUE + SQuAD) after fine-tuning. We pretrain and finetune our models with 16 TPU-v3 chips with data parallelism. All large models have a model parallelism of $2$ and XL models have a model parallelism of $8$.

 \paragraph{Model Sizes} We consider several different model sizes for each architecture. For models that are straightforward to scale, we simply follow the standard convention in \citet{raffel2019exploring}, moving from small to base, to large and XL. We include a tiny version of each model to observe how different models behave at lower compute regions. For models where it was not straightforward to scale (e.g., Universal Transformers, ALBERT), we tried to scale them in a similar fashion but faced obvious limitations such as getting ALBERT to have the same number of parameters as T5 XL without incurring a huge number of cost in terms of FLOPs. For convolutional models, we consider $d_{\rm{model}}$ to be the hidden size (i.e., channel depth) for the one-dimensional convolution layers. Values such as $d_{\rm{kv}}, N_H$ then become redundant. Details on scaling details\footnote{The largest Switch transformer was scaled in a pretty sub-optimal way. So we don't think it is representative of the full potential of the Switch family. Take the last data point of Switch with a pinch of salt. } of each architecture can be found in the supplementary material.

\begin{figure*}[t]
\small
     \centering
    \begin{subfigure}[b]{0.23\textwidth}
         \centering
         \includegraphics[width=\textwidth]{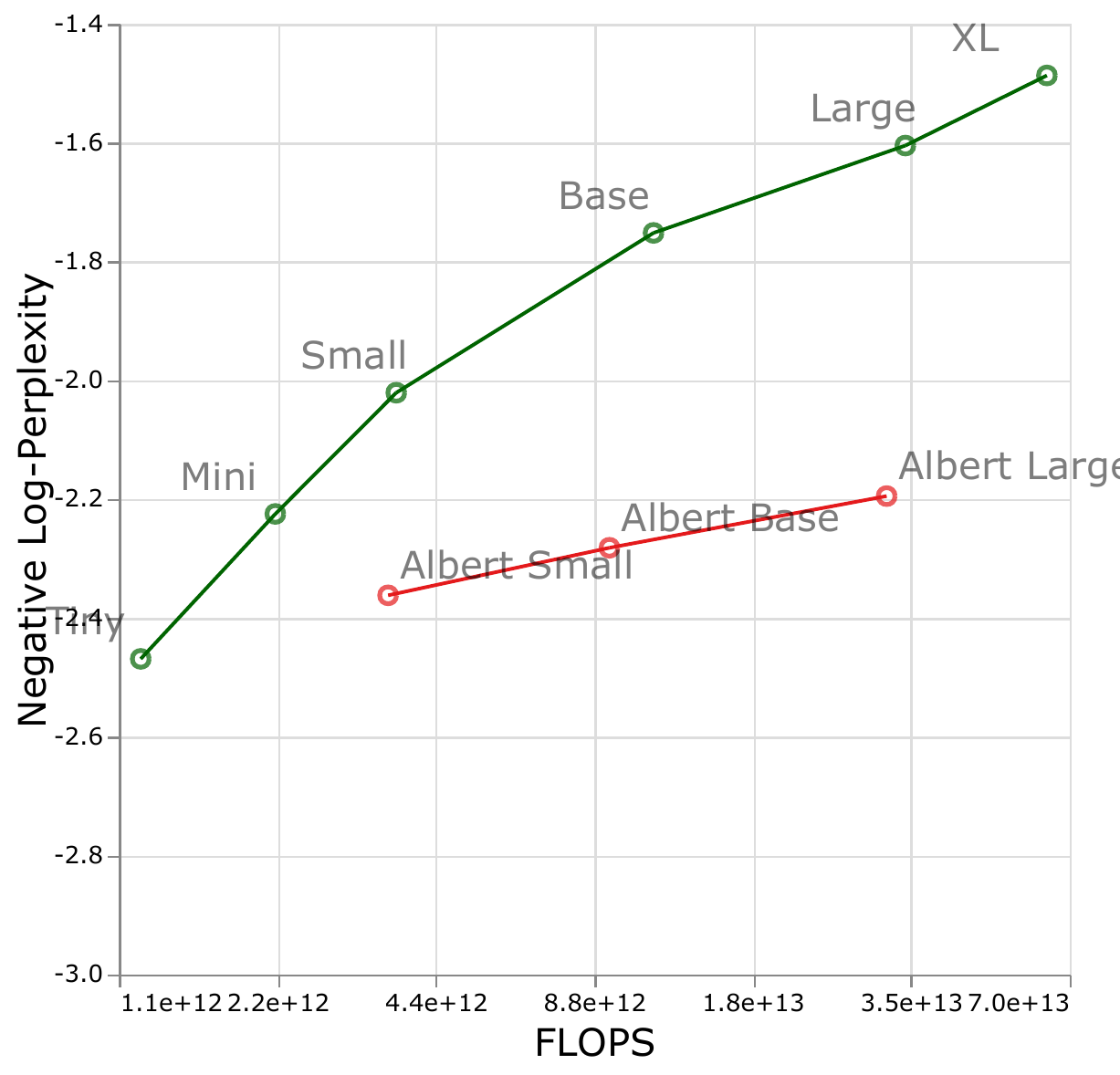}
         \caption{ALBERT}
         \label{fig:albert_us}
     \vspace{5pt} \end{subfigure}
    \begin{subfigure}[b]{0.25\textwidth}
         \centering
         \includegraphics[width=\textwidth]{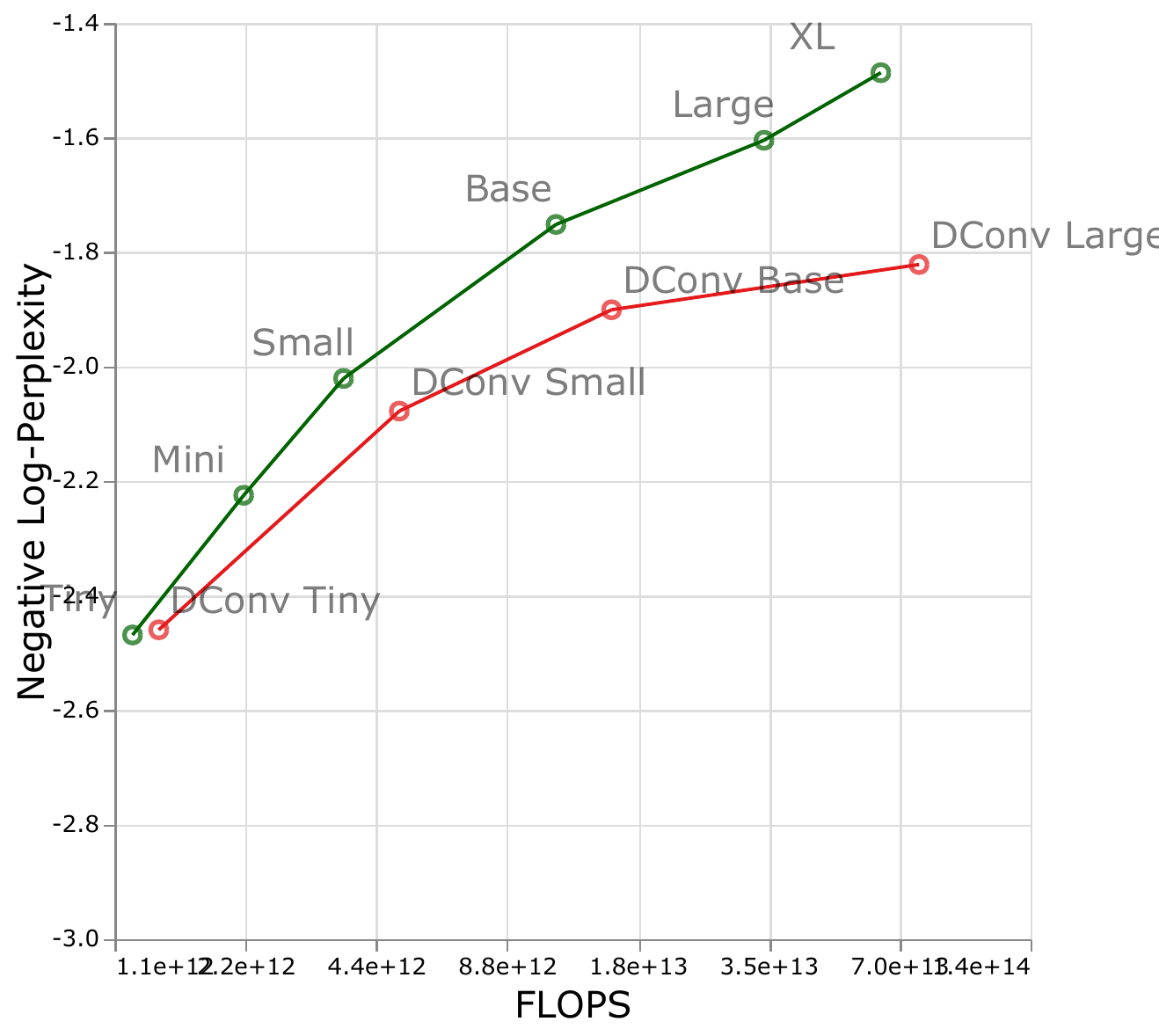}
         \caption{DConv}
         \label{fig:dconv_us}
     \vspace{5pt} \end{subfigure}
     \begin{subfigure}[b]{0.23\textwidth}
         \centering
         \includegraphics[width=\textwidth]{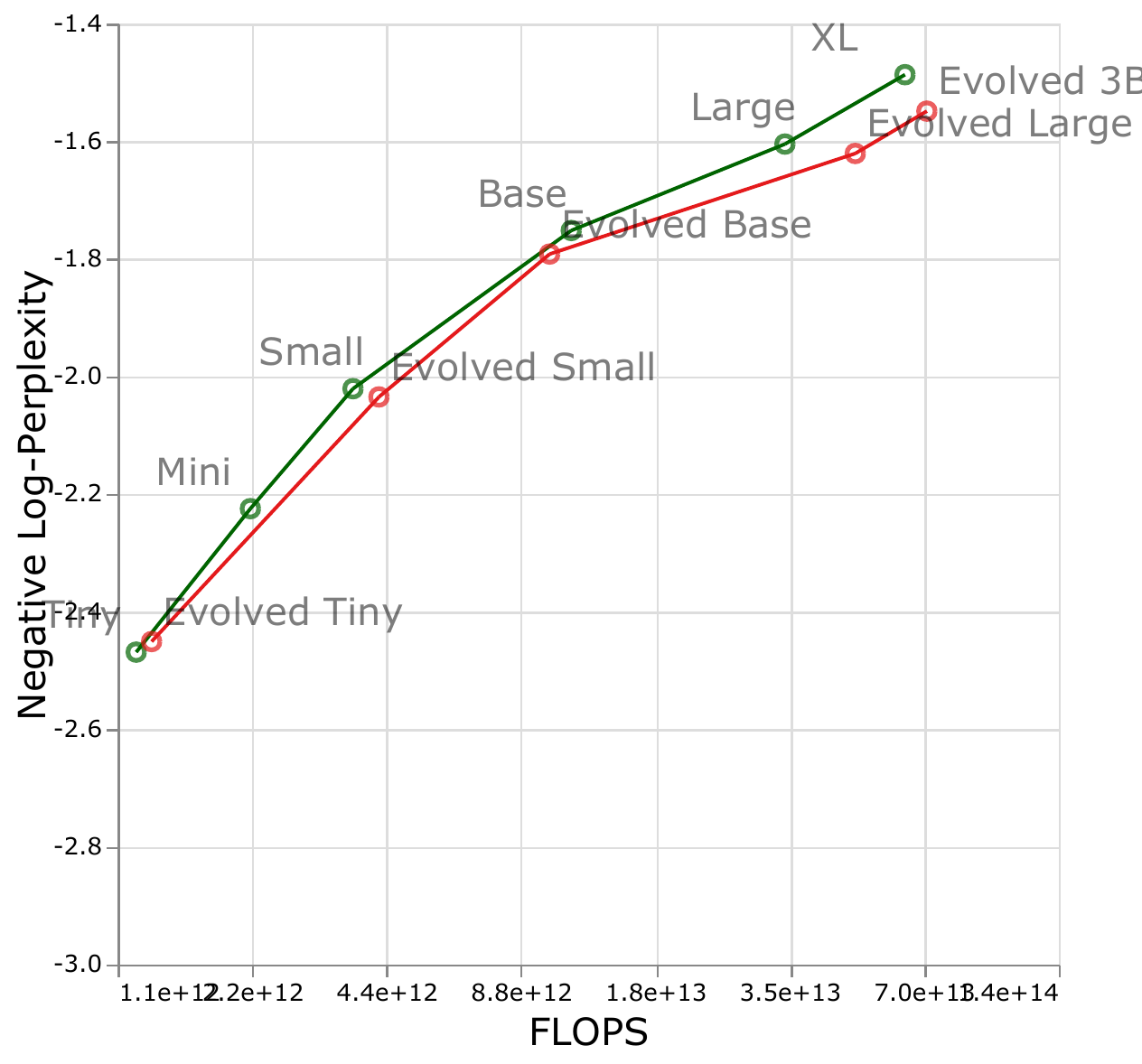}
         \caption{Evolved}
         \label{fig:evolved_us}
     \vspace{5pt} \end{subfigure}
     \begin{subfigure}[b]{0.23\textwidth}
         \centering
         \includegraphics[width=\textwidth]{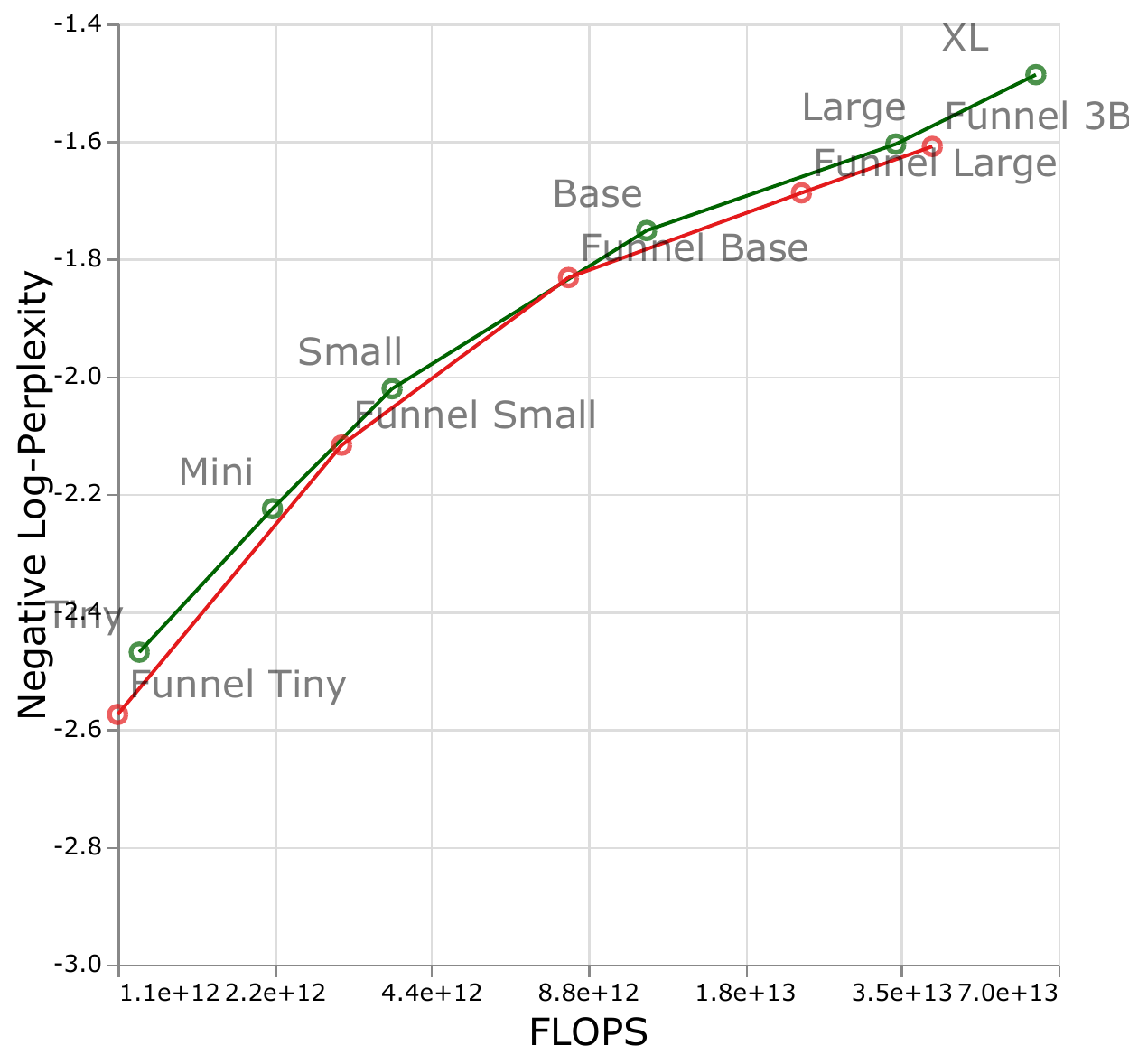}
         \caption{Funnel}
         \label{fig:funnel_us}
     \vspace{5pt} \end{subfigure}
     \begin{subfigure}[b]{0.23\textwidth}
         \centering
         \includegraphics[width=\textwidth]{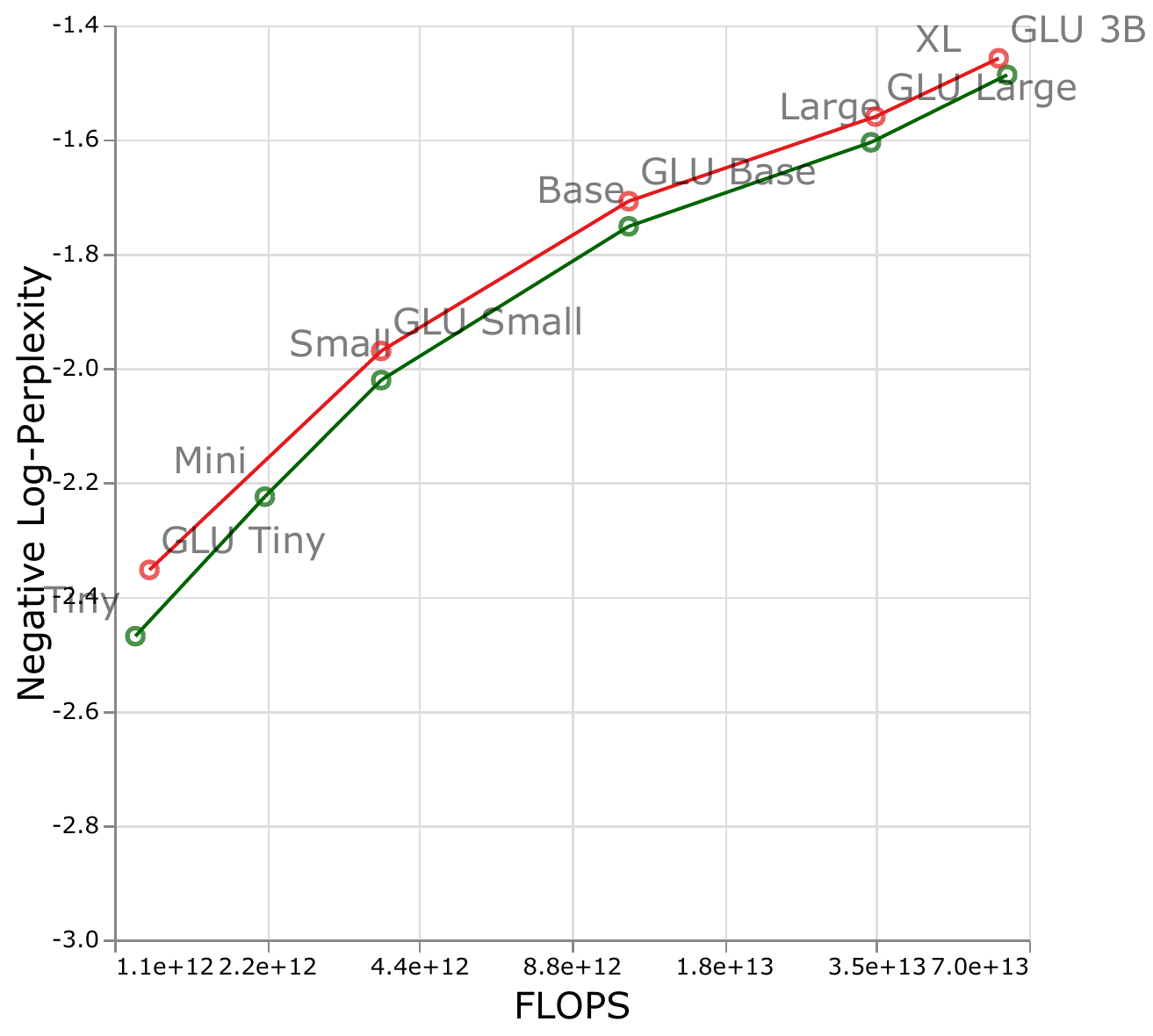}
         \caption{Transformer-GLU}
         \label{fig:glu_us}
     \vspace{5pt} \end{subfigure}
     \begin{subfigure}[b]{0.23\textwidth}
         \centering
         \includegraphics[width=\textwidth]{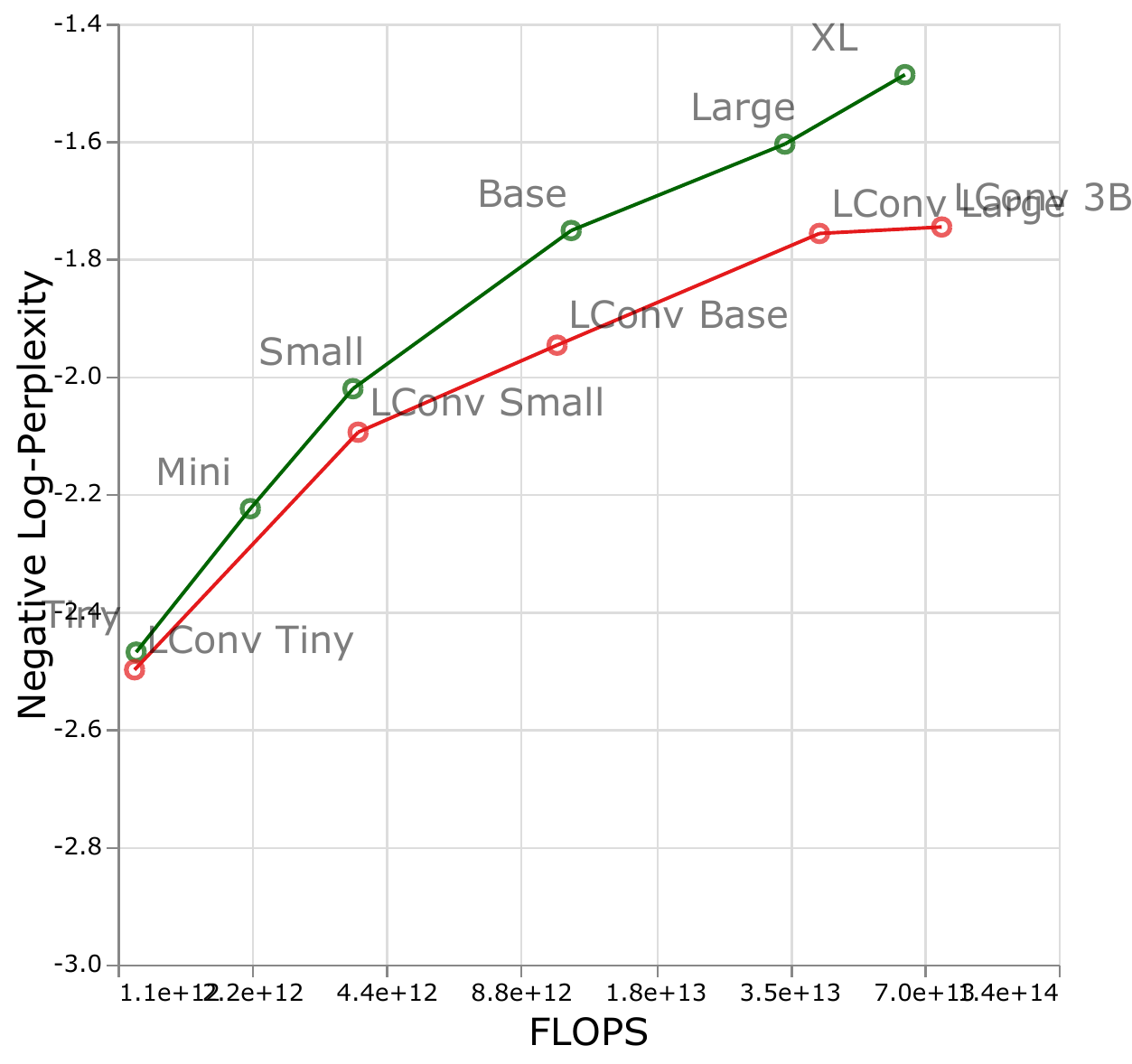}
         \caption{LConv}
         \label{fig:lconv_us}
     \vspace{5pt} \end{subfigure}
        \begin{subfigure}[b]{0.25\textwidth}
         \centering
         \includegraphics[width=\textwidth]{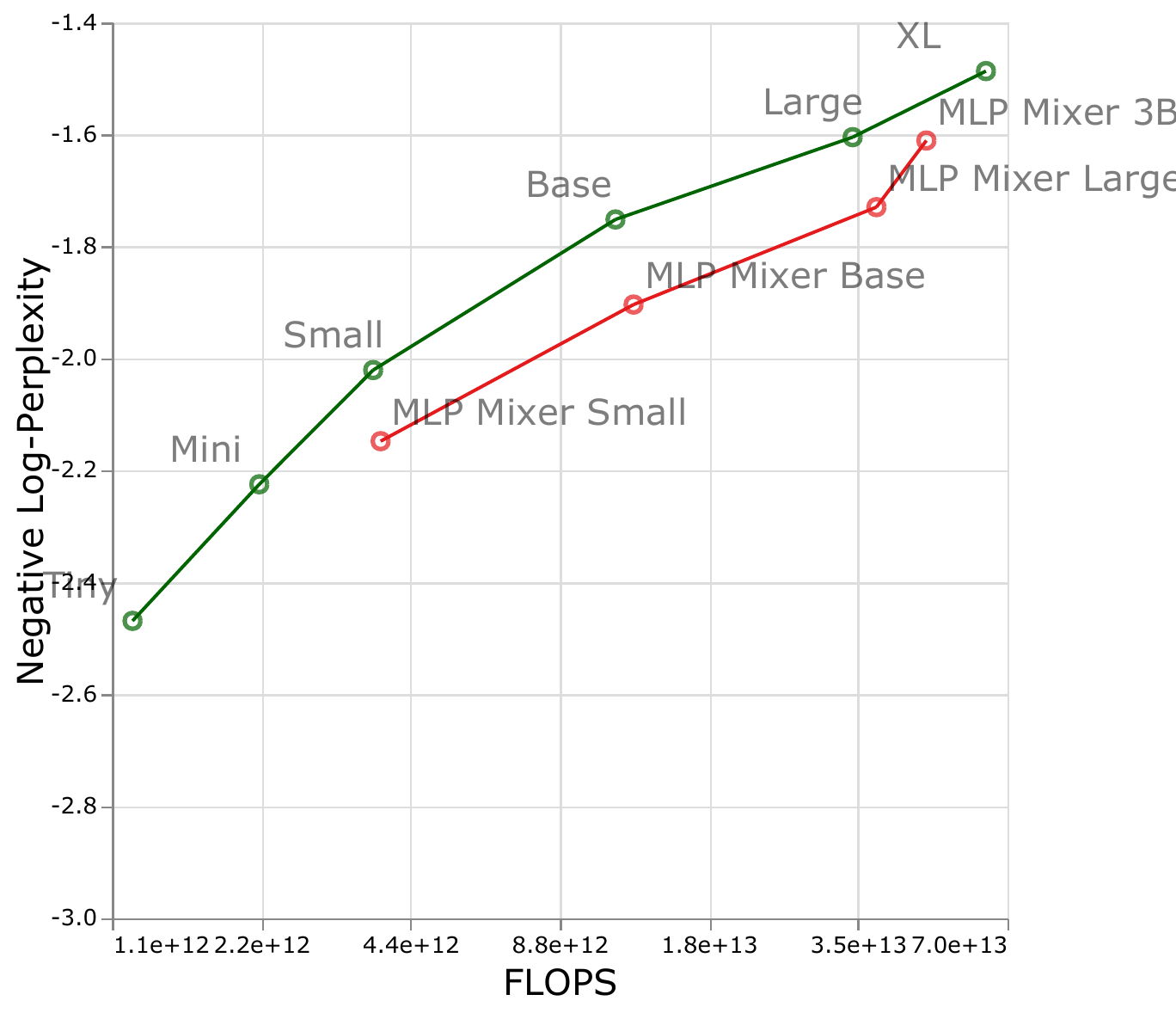}
         \caption{MLP Mixer}
         \label{fig:mixer_us}
     \vspace{5pt} \end{subfigure}
     \begin{subfigure}[b]{0.24\textwidth}
         \centering
         \includegraphics[width=\textwidth]{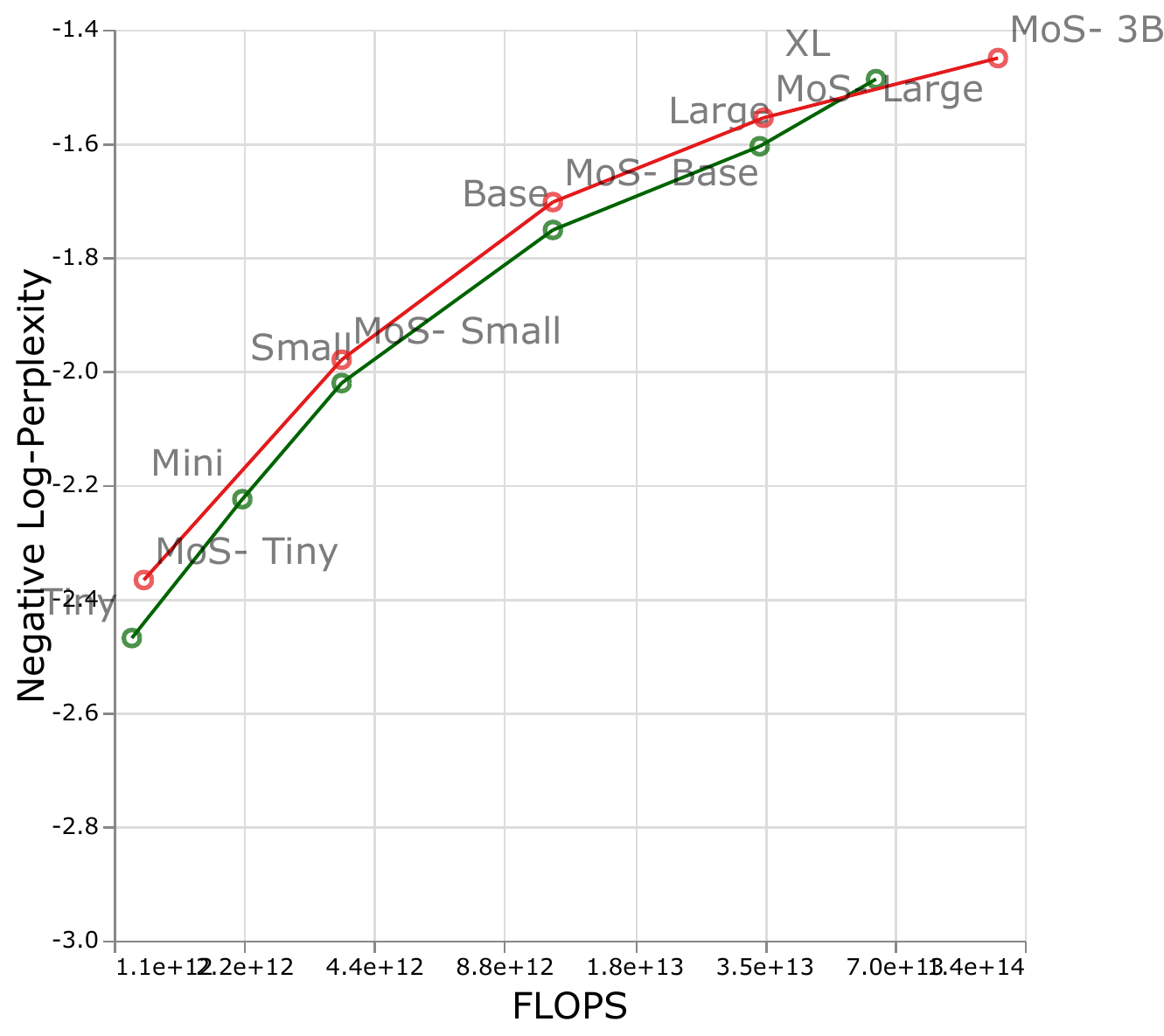}
         \caption{MoS Transformer}
         \label{fig:mos_us}
     \vspace{5pt} \end{subfigure}
    \begin{subfigure}[b]{0.23\textwidth}
         \centering
         \includegraphics[width=\textwidth]{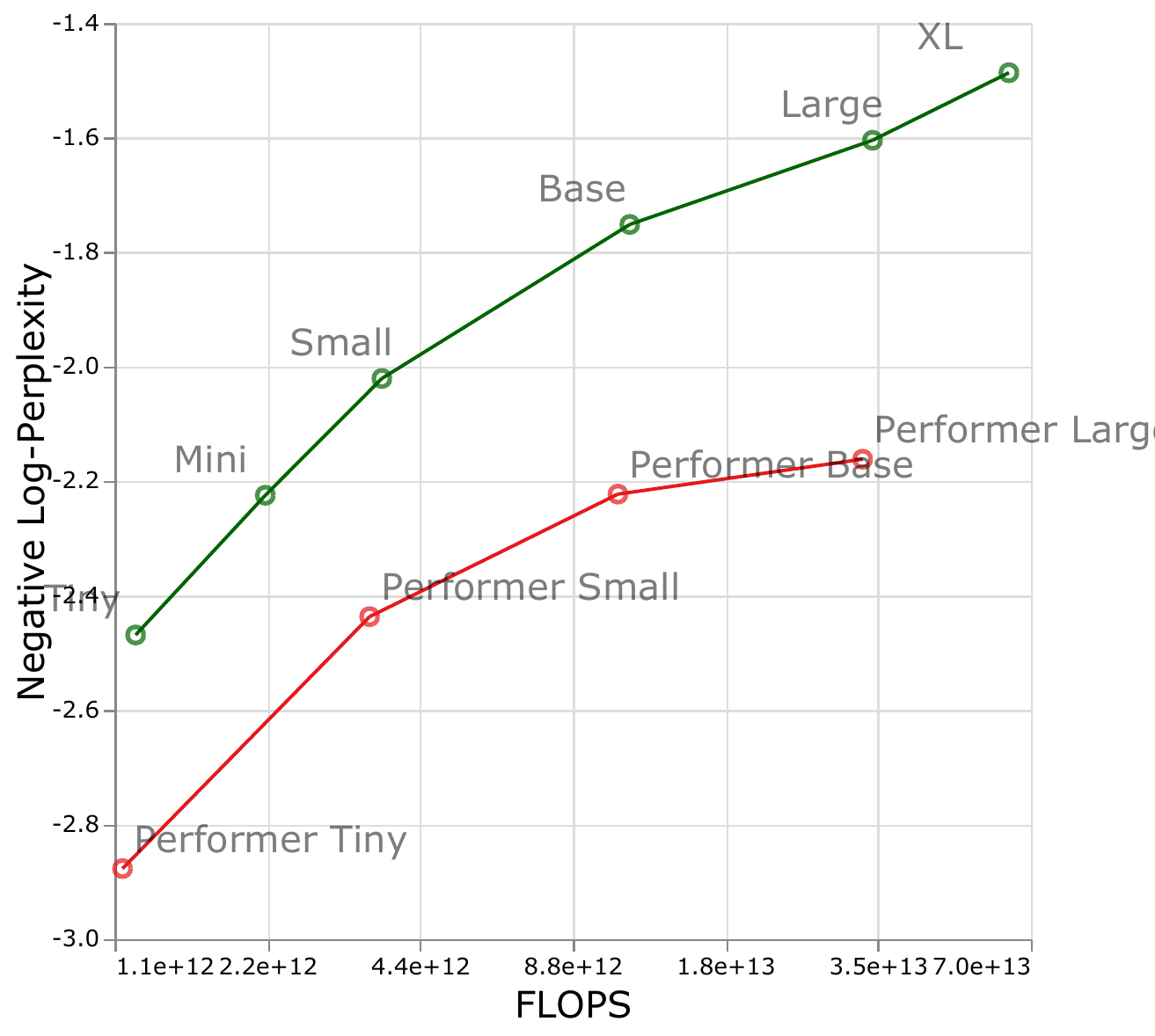}
         \caption{Performer}
         \label{fig:performer_us}
    \vspace{5pt} \end{subfigure}
    \begin{subfigure}[b]{0.23\textwidth}
         \centering
         \includegraphics[width=\textwidth]{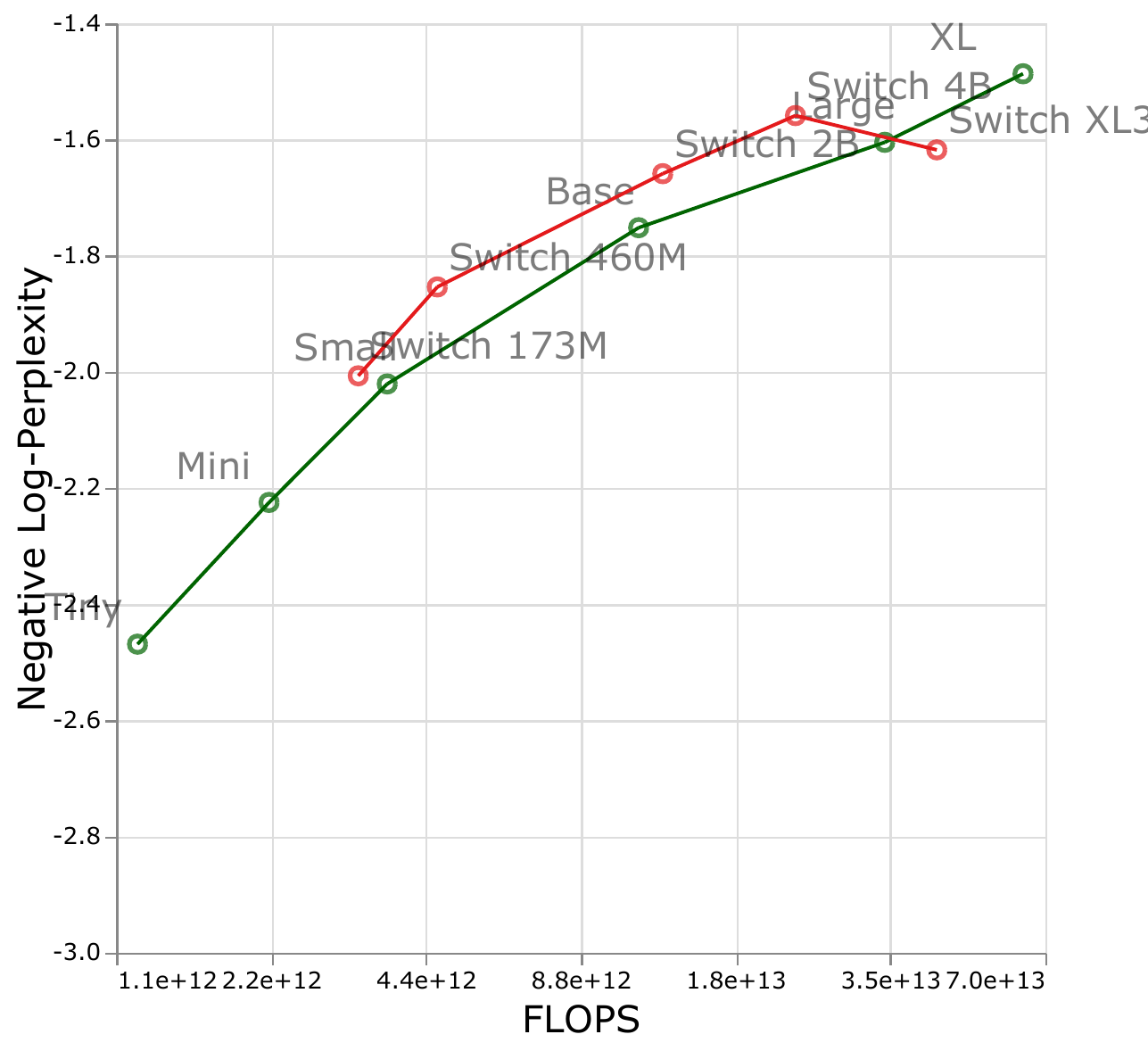}
         \caption{Switch Transformer}
         \label{fig:switch_us}
     \vspace{5pt} \end{subfigure}
    \begin{subfigure}[b]{0.24\textwidth}
         \centering
         \includegraphics[width=\textwidth]{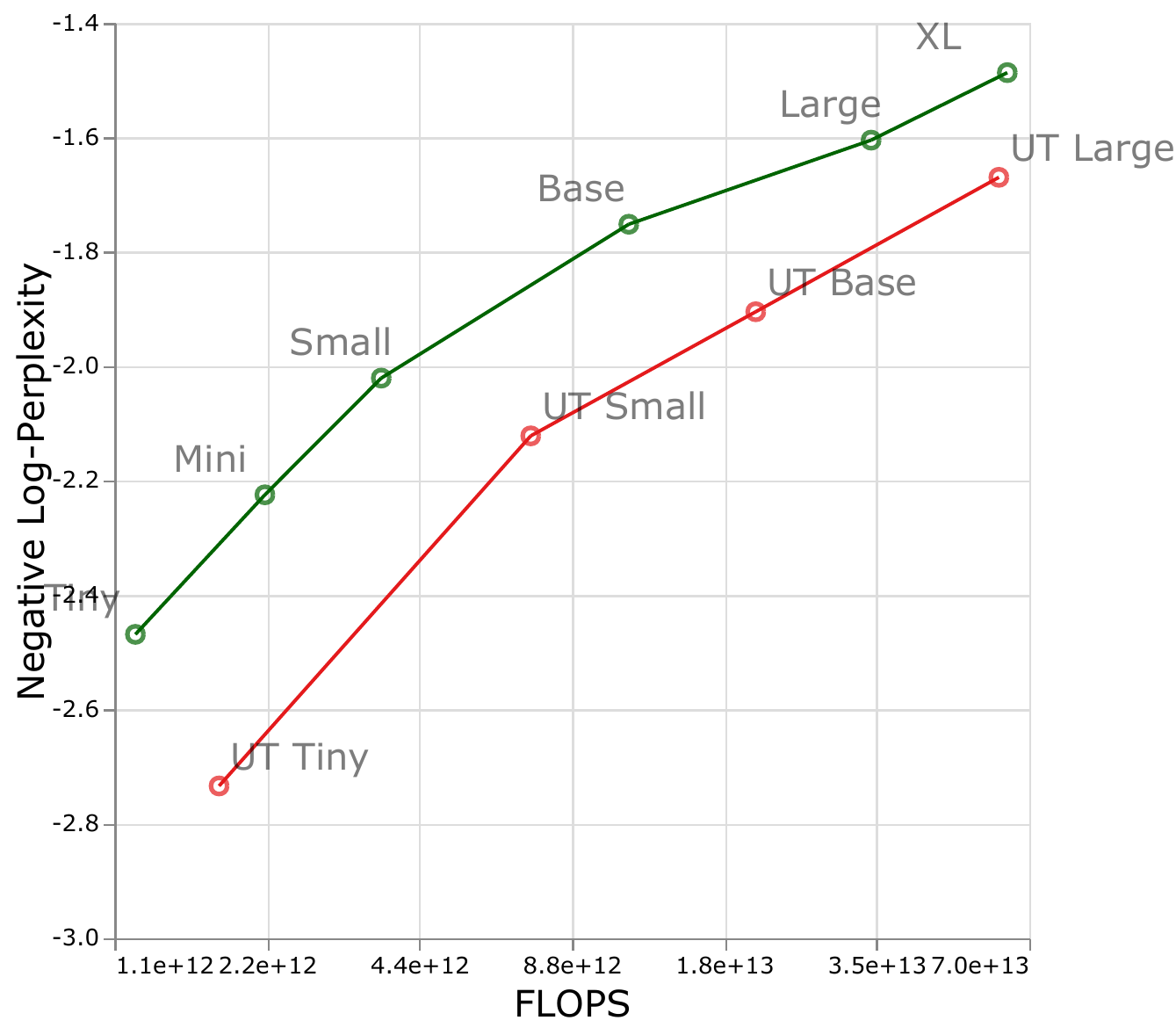}
         \caption{Universal Transformer}
         \label{fig:ut_us}
     \vspace{5pt} \end{subfigure}
    \caption{Upstream Negative Log-Perplexity of vanilla Transformer compared to other models.}
    \label{fig:upstreams_scaling}
\end{figure*}
 
 \subsection{Main Results}
 We report the main results of this paper in Table \ref{tab:all_results}. We report the number of trainable parameters, FLOPs (of a single forward pass) and speed (steps per second). We also report on validation perplexity (on upstream pre-training) and results on 17 downstream tasks. The results are reported aggregates of GLUE, SuperGLUE and SQuAD. While we use the same Mesh TensorFlow-based codebase used by~\citet{raffel2019exploring} and hence expect our experimental results to match theirs, we verify that our T5 base does achieve similar results to what is reported in \citet{raffel2019exploring}.

 \subsection{Do all models scale the same way?}
 This section investigates if all model architectures scale in the same way. 

\paragraph{Upstream Perplexity}

Figure \ref{fig:upstreams_scaling} reports the scaling behaviour of all models as we increase the number of FLOPs. We observe that the scaling behaviour of all models are quite unique and distinct, i.e., most of them are quite different from standard Transformers. Perhaps the biggest finding here is that most models (e.g., LConv, Evolved) all seem to be on-par or better than standard Transformers but fail to scale with a higher compute budget.
Another interesting trend is that ``linear" Transformers such as Performer fail to scale as shown in Figure~\ref{fig:performer_us}. The pre-training perplexity metric only decreases by 2.7\% going from base to large scale compared to 8.4\% of the vanilla Transformer.

\paragraph{Downstream Transfer}

\begin{figure*}[t]
\small
     \centering
    \begin{subfigure}[b]{0.23\textwidth}
         \centering
         \includegraphics[width=\textwidth]{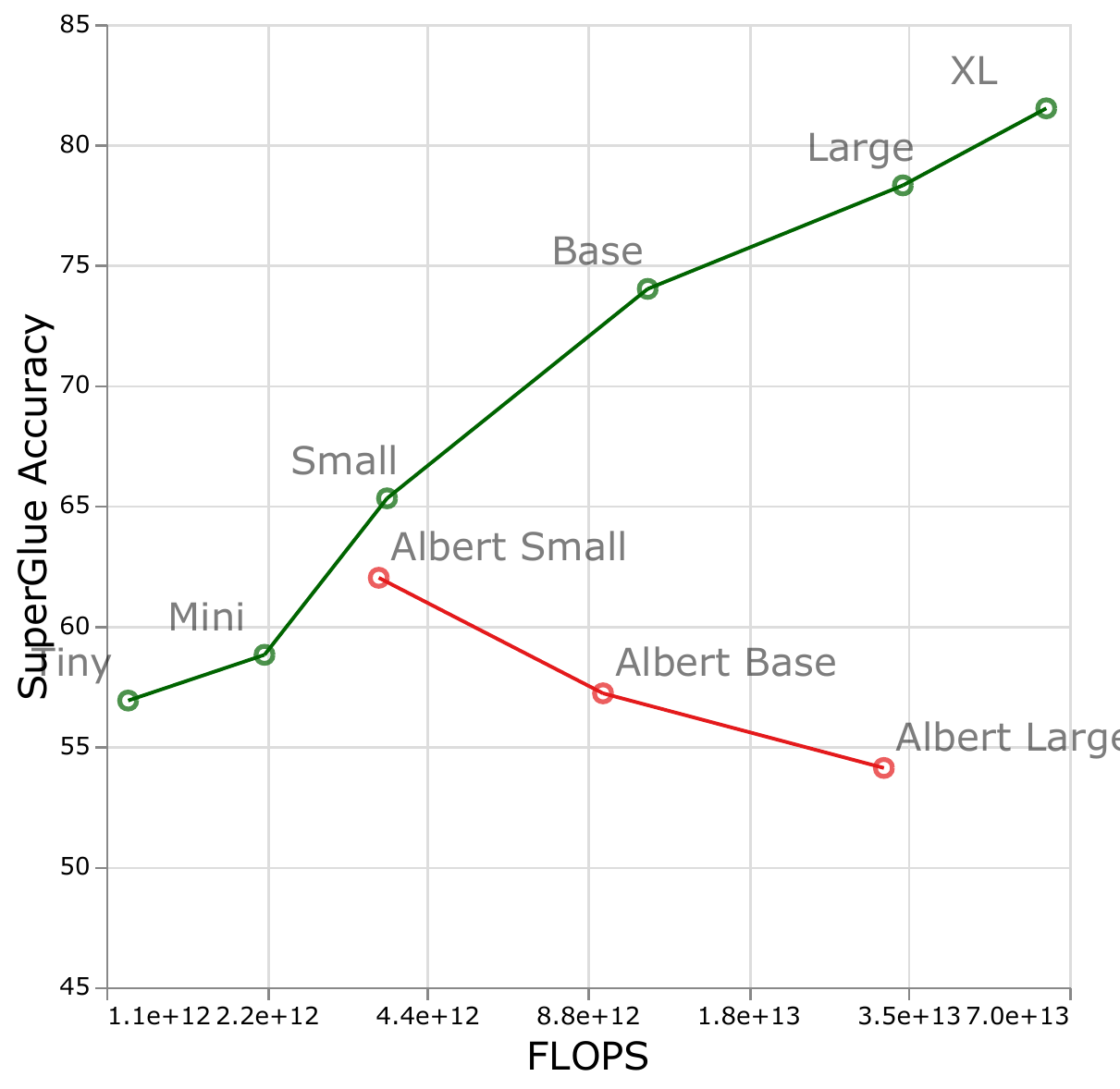}
         \caption{ALBERT}
         \label{fig:albert_ds}
     \vspace{5pt} \end{subfigure}
    \begin{subfigure}[b]{0.25\textwidth}
         \centering
         \includegraphics[width=\textwidth]{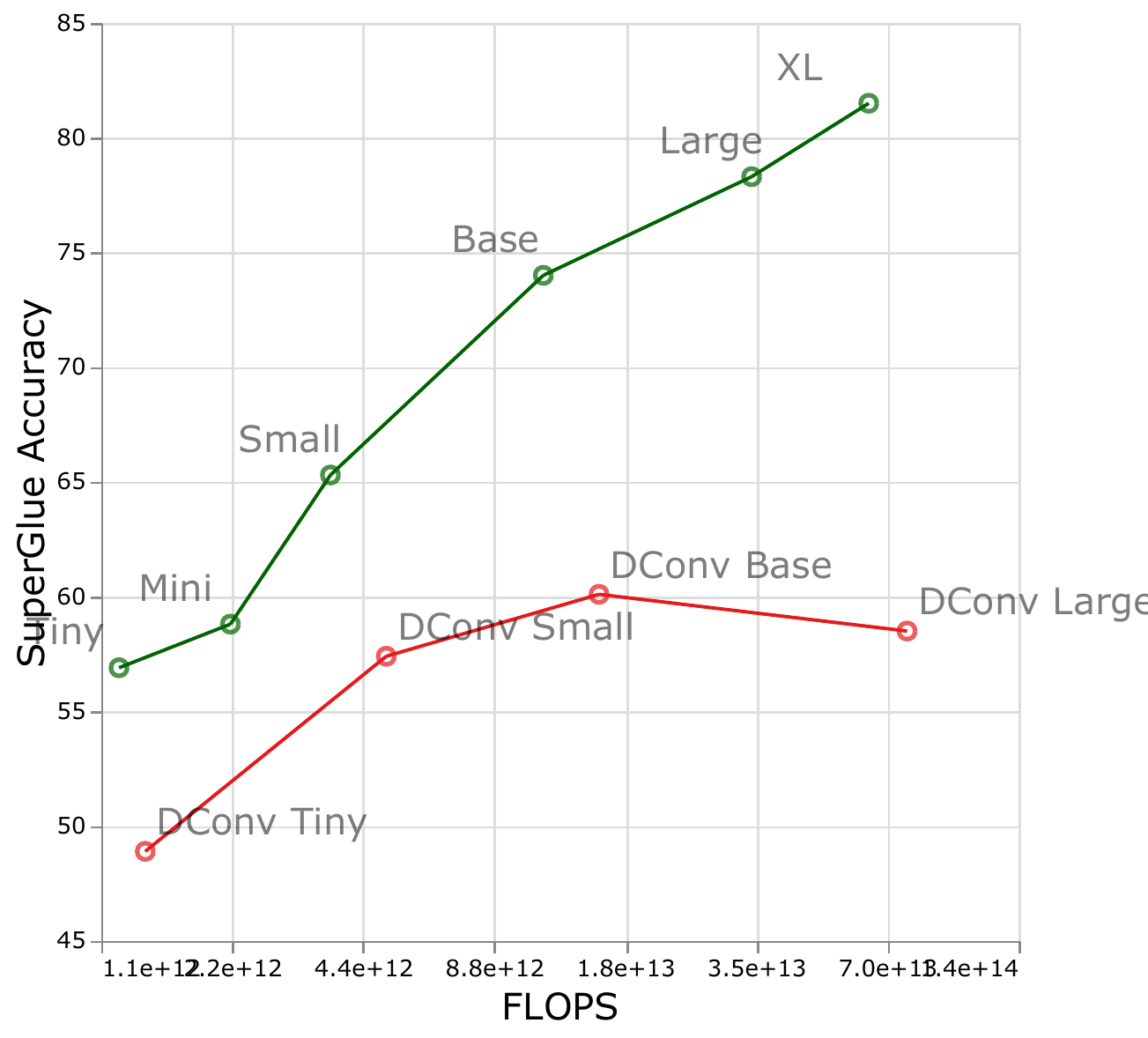}
         \caption{DConv}
         \label{fig:dconv_ds}
     \vspace{5pt} \end{subfigure}
     \begin{subfigure}[b]{0.23\textwidth}
         \centering
         \includegraphics[width=\textwidth]{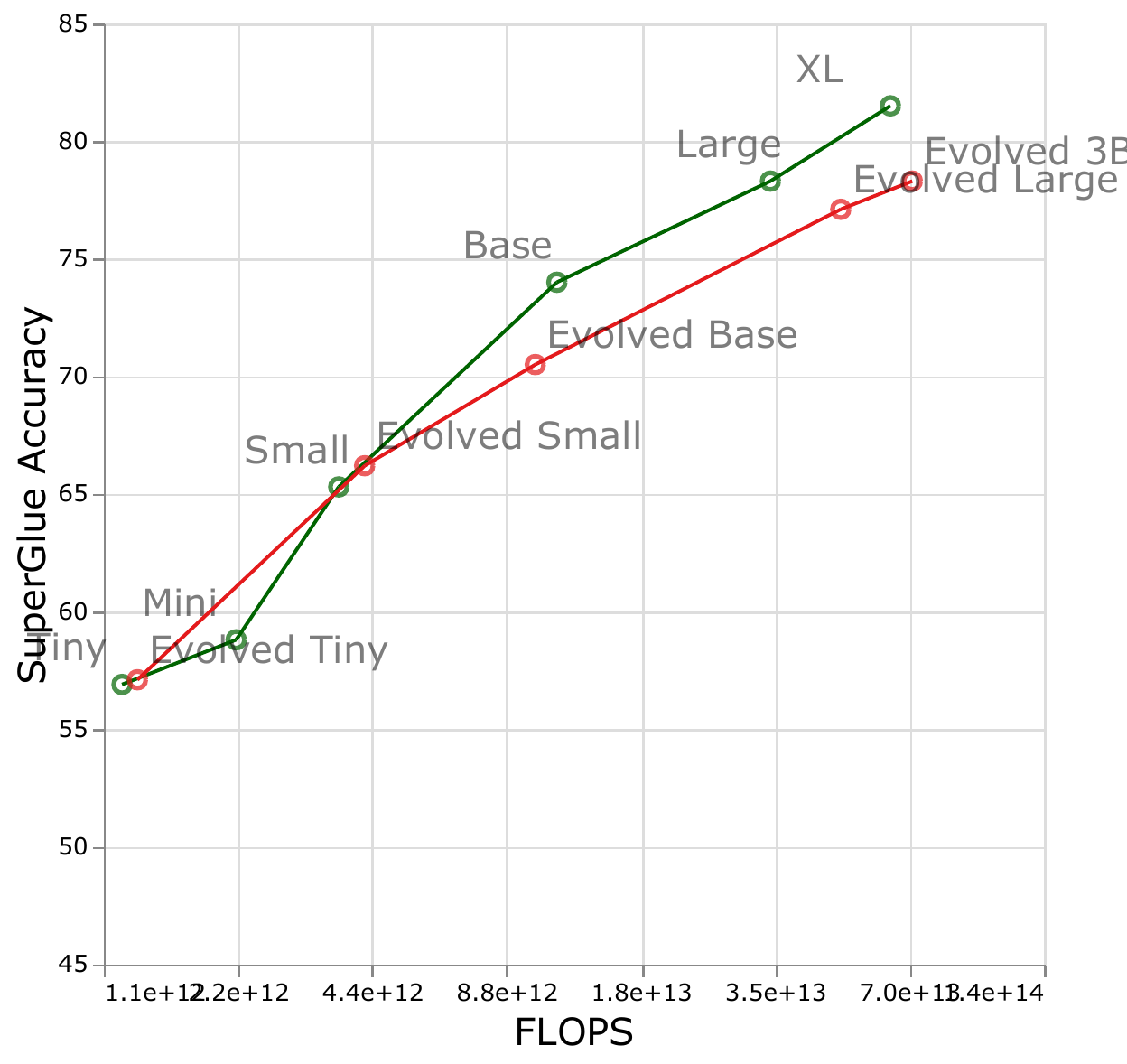}
         \caption{Evolved}
         \label{fig:evolved_ds}
     \vspace{5pt} \end{subfigure}
     \begin{subfigure}[b]{0.23\textwidth}
         \centering
         \includegraphics[width=\textwidth]{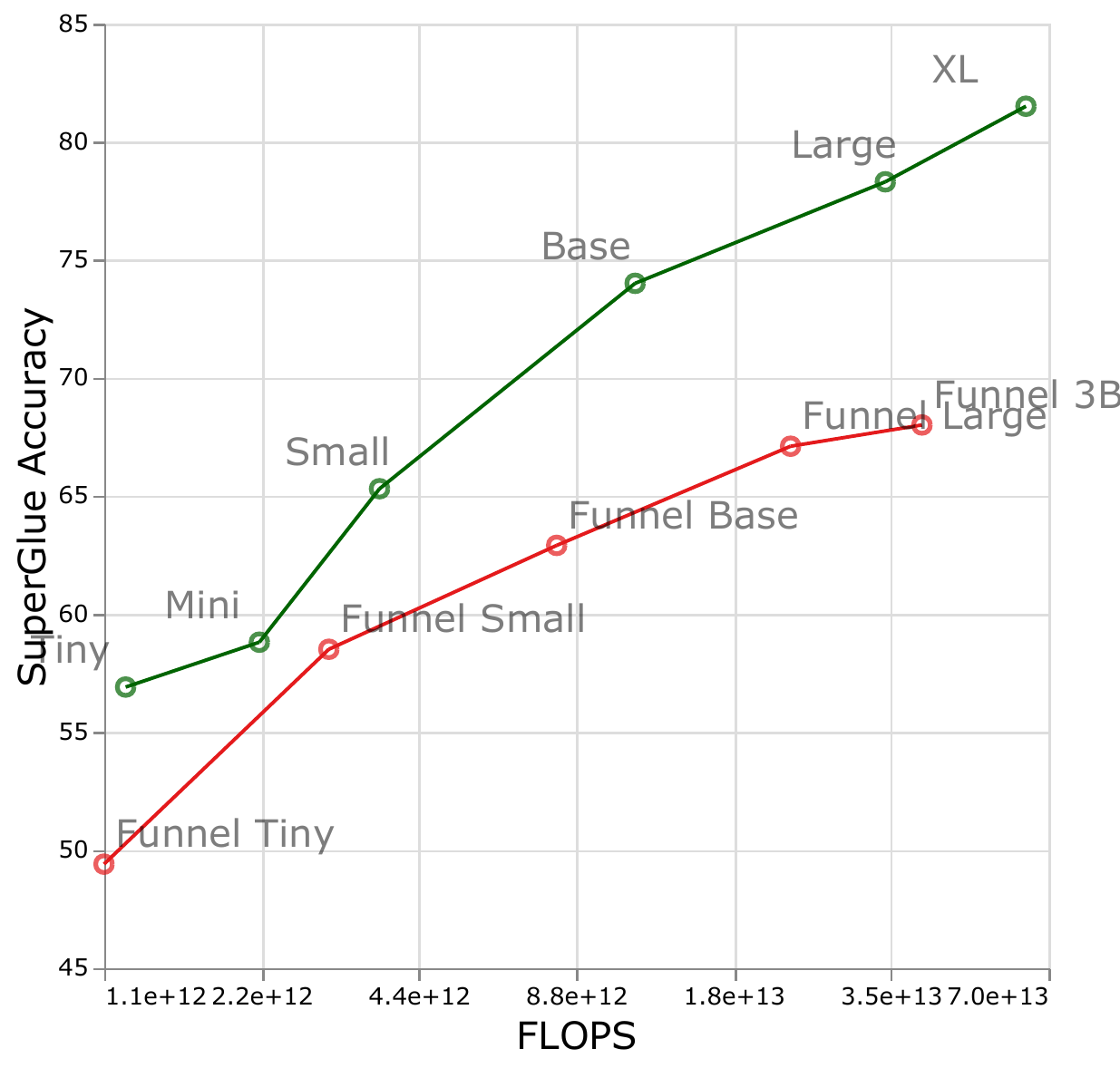}
         \caption{Funnel}
         \label{fig:funnel_ds}
     \vspace{5pt} \end{subfigure}
     \begin{subfigure}[b]{0.23\textwidth}
         \centering
         \includegraphics[width=\textwidth]{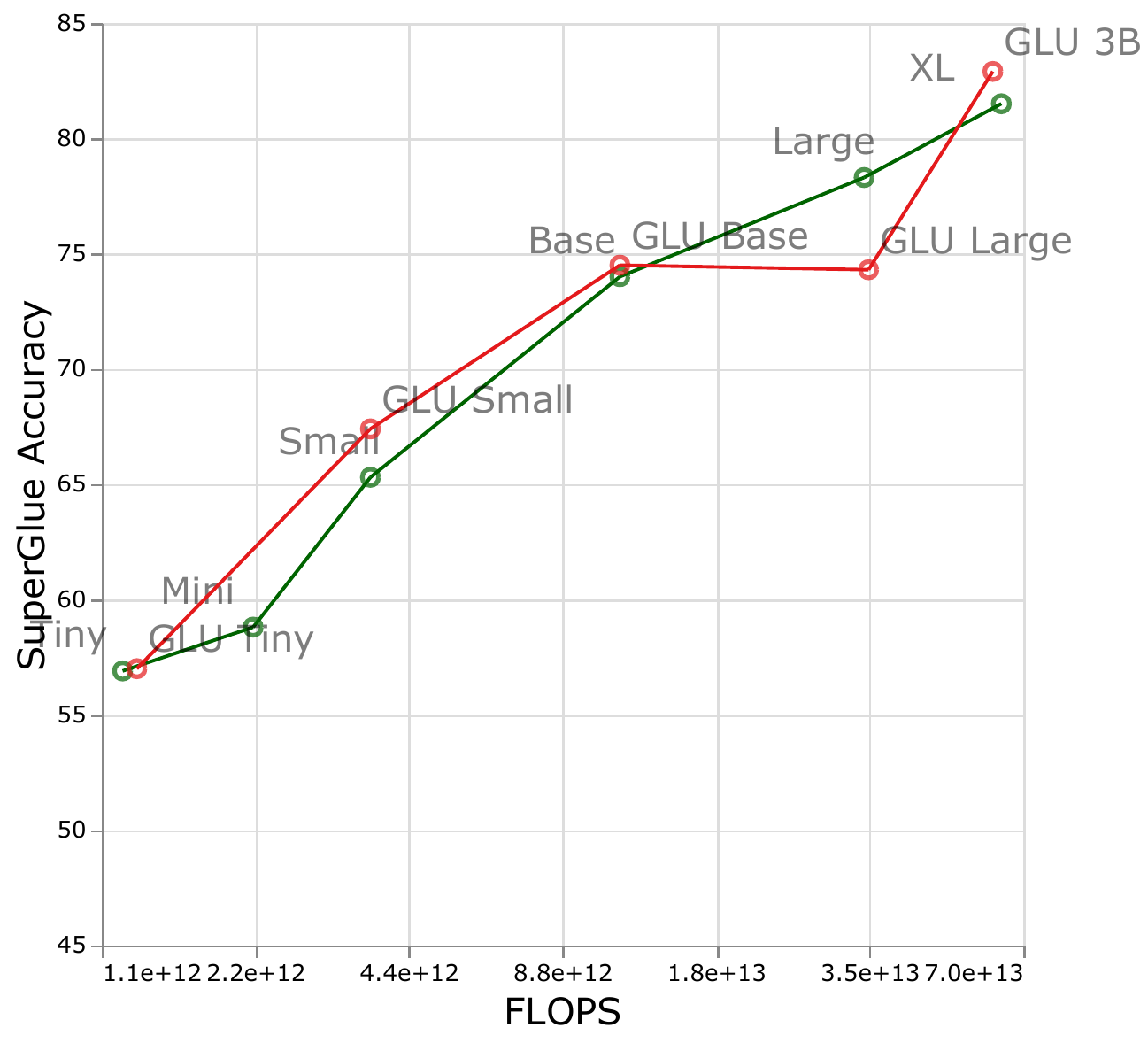}
         \caption{Transformer-GLU}
         \label{fig:glu_ds}
     \vspace{5pt} \end{subfigure}
     \begin{subfigure}[b]{0.23\textwidth}
         \centering
         \includegraphics[width=\textwidth]{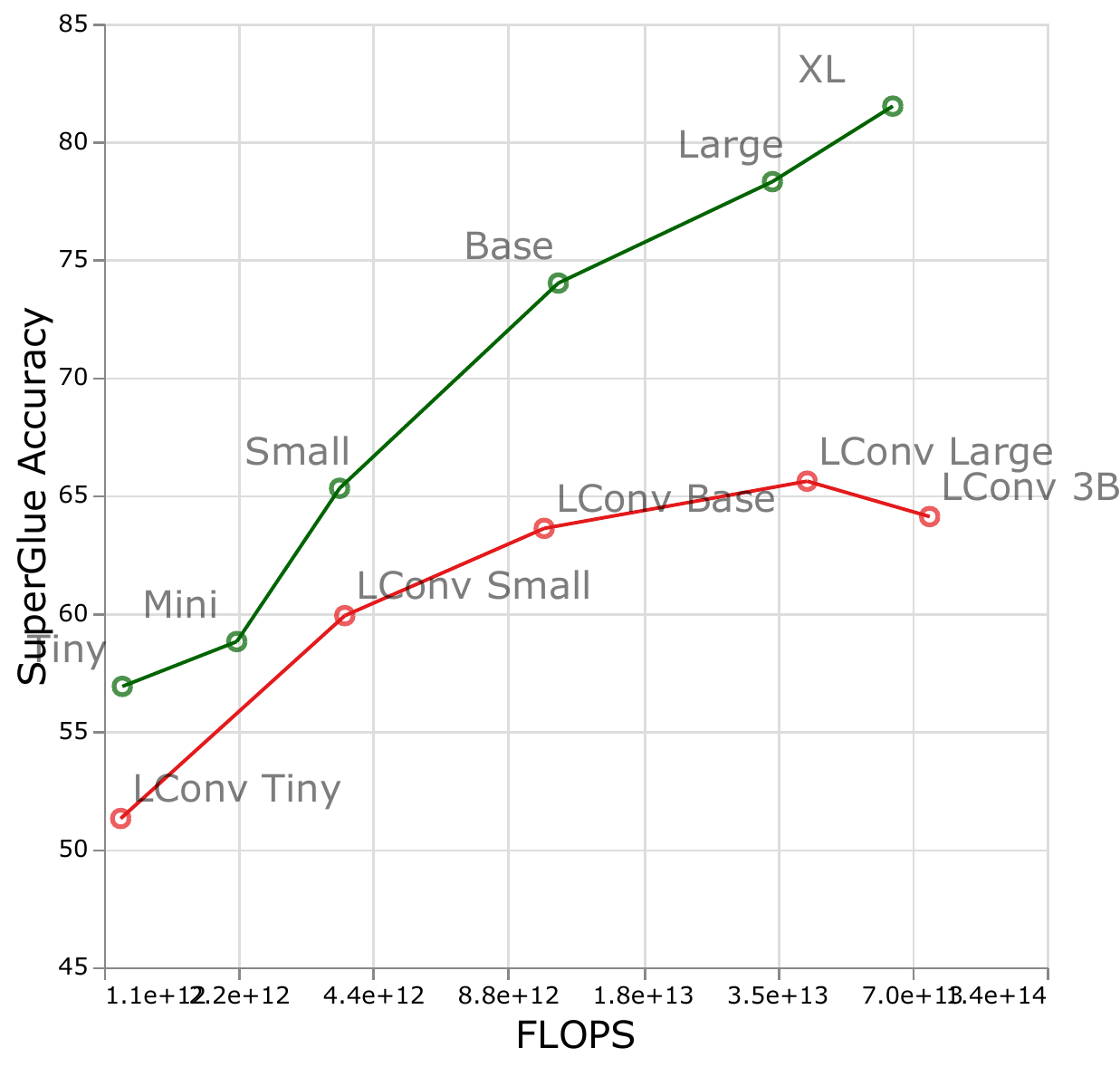}
         \caption{LConv}
         \label{fig:lconv_ds}
     \vspace{5pt} \end{subfigure}
        \begin{subfigure}[b]{0.25\textwidth}
         \centering
         \includegraphics[width=\textwidth]{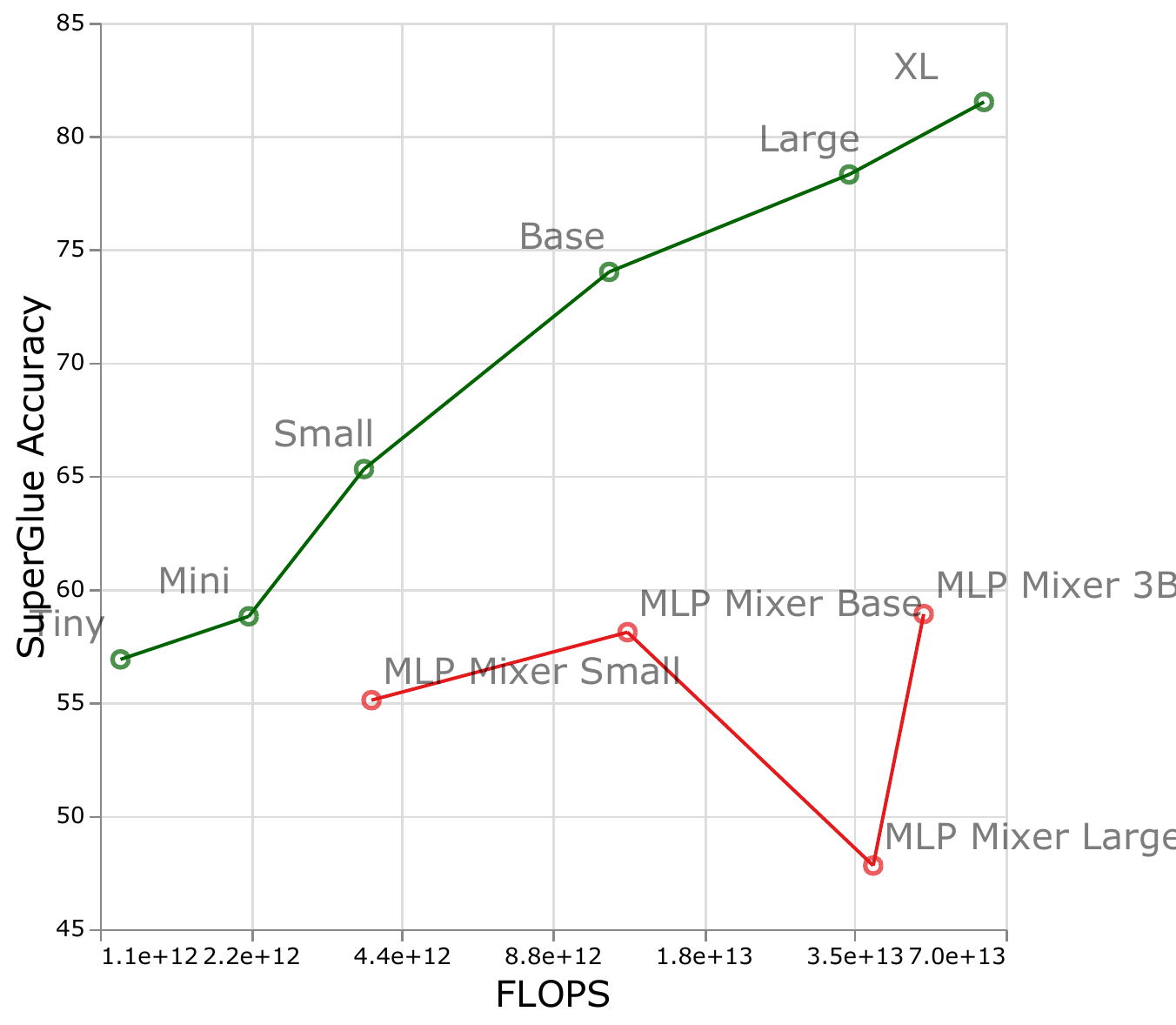}
         \caption{MLP Mixer}
         \label{fig:mixer_ds}
     \vspace{5pt} \end{subfigure}
     \begin{subfigure}[b]{0.24\textwidth}
         \centering
         \includegraphics[width=\textwidth]{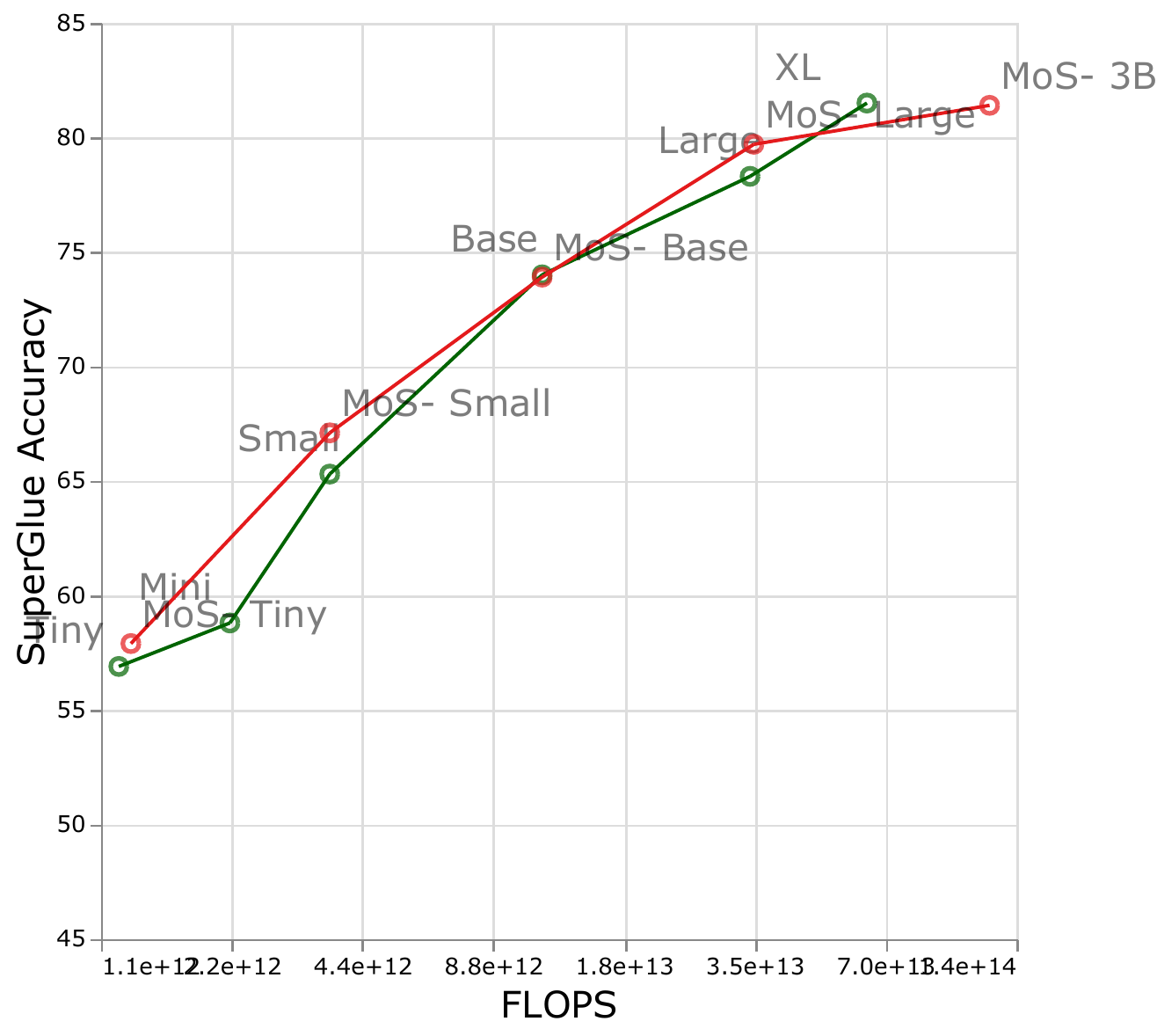}
         \caption{MoS Transformer}
         \label{fig:mos_ds}
     \vspace{5pt} \end{subfigure}
    \begin{subfigure}[b]{0.23\textwidth}
         \centering
         \includegraphics[width=\textwidth]{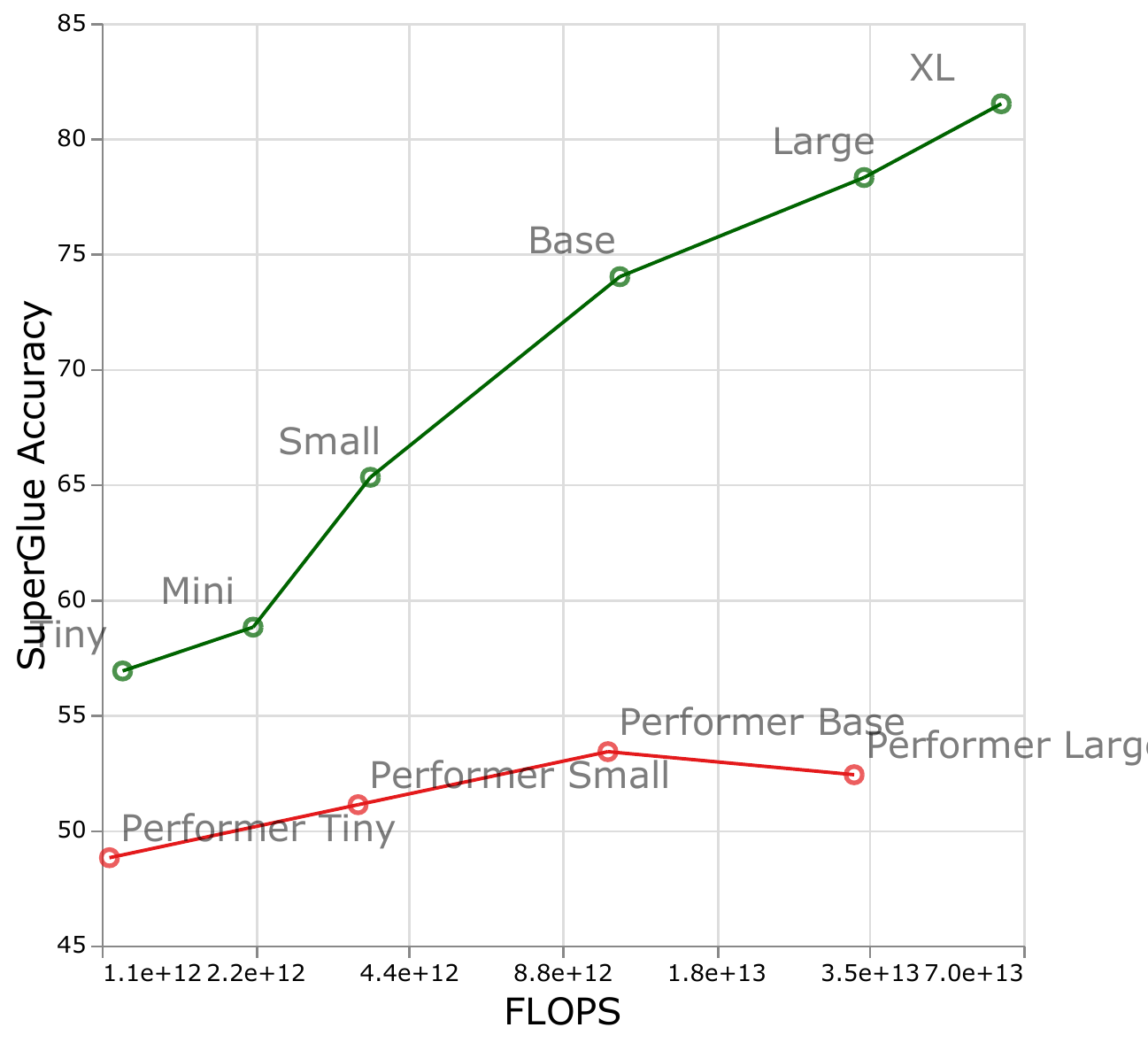}
         \caption{Performer}
         \label{fig:performer_ds}
    \vspace{5pt} \end{subfigure}
    \begin{subfigure}[b]{0.23\textwidth}
         \centering
         \includegraphics[width=\textwidth]{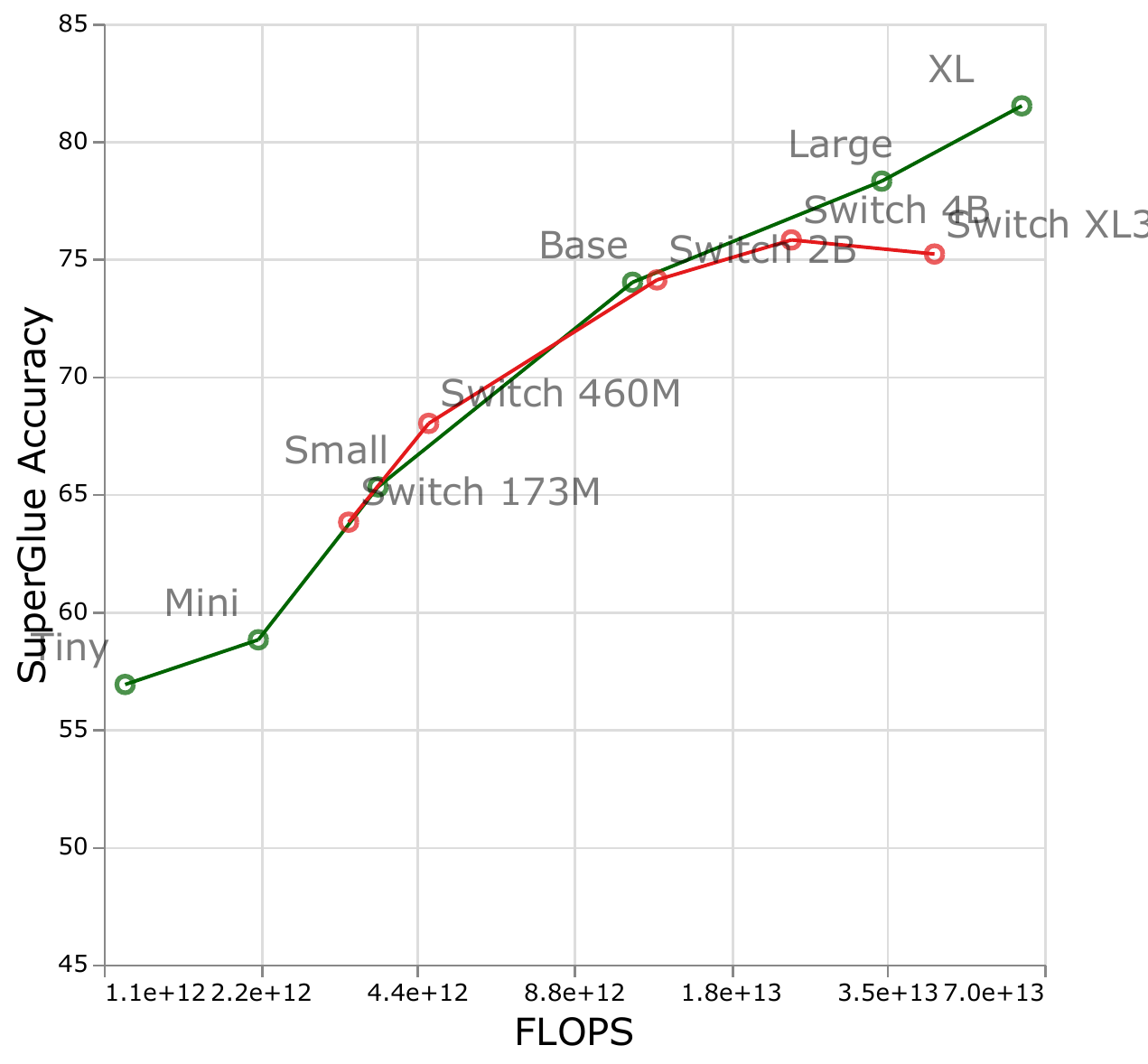}
         \caption{Switch Transformer}
         \label{fig:switch_ds}
     \vspace{5pt} \end{subfigure}
    \begin{subfigure}[b]{0.24\textwidth}
         \centering
         \includegraphics[width=\textwidth]{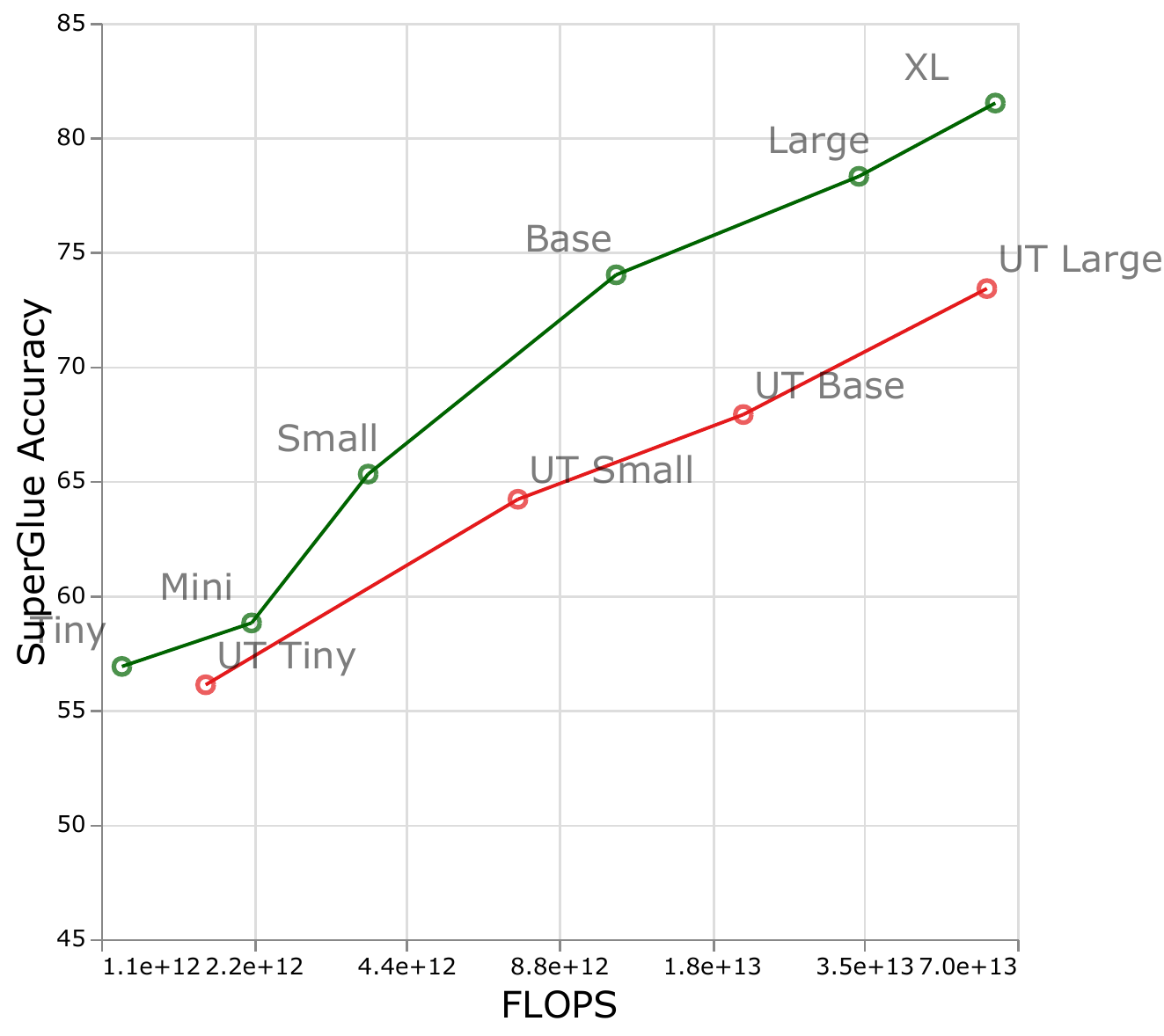}
         \caption{Universal Transformer}
         \label{fig:ut_ds}
     \vspace{5pt} \end{subfigure}
    \caption{Downstream accuracy of vanilla Transformer compared to other models.}
    \label{fig:downstream_scaling}
\end{figure*}
Figure \ref{fig:downstream_scaling} reports the scaling curves of all models on downstream transfer. The overall finding that most models have distinct scaling curves compared to Transformers is also evident in downstream tasks. It is also noteworthy that most models have a different upstream and downstream scaling curve. We find that some models such as Funnel Transformer and LConvs that seem to hold out pretty well on upstream but suffer substantially on downstream. As for Performer, the performance (disparity) seems to be even greater in downstream as compared to upstream. Notably, the SuperGLUE downstream tasks generally require pseudo cross-attention on the encoder, which models such as convolutions are not equipped to handle \citep{tay2021pre}. To this end, we find that certain models may have difficulty learning the downstream tasks despite good upstream performance. 

\subsection{Are the best models at each scale different?}
Figure \ref{fig:overview} shows the Pareto-frontier when plotting compute against upstream and downstream performance. Since the colors of the plot represent different models, we can observe that the best model for every scale and compute region might be different. Moreover, from Figure \ref{fig:downstream_scaling}, we can also observe this. For example, the Evolved Transformer seems to do well against the standard Transformer at tiny to small region (downstream) but this quickly changes when scaling the model up. We also observe this with MoS-Transformer where it clearly outperforms vanilla Transformers at some regions but not at others.

 \begin{table*}[]
     \centering
     \small
       \caption{Results on pre-training and finetuning ten different model architectures. Full results (further varying hyperparameters of these models) can be found in the Appendix.}
    \resizebox{\textwidth}{!}{%
     \begin{tabular}{l|ccc|cccc}
     \hline
        Model   & \#Params & FLOPs & Speed & Neg Log Ppl & GLUE & SGLUE & SQuAD  \\
        \hline
       Transformer Tiny   &  16M & 1.21 & 38.4 &  -2.47 & 69.3	& 56.9 &	73.6 \\
       Transformer Small & 60M & 3.70 & 22.7& -2.02 & 78.1 &	65.3 &	81.9 \\ 
       Transformer Base & 223M & 11.4 &9.3  & -1.75 &  83.8 &	74.0 &	86.3\\ 
       Transformer Large & 738M & 34.3 & 3.6 &  -1.61 & 86.4 &	78.3 &	88.6\\
       Transformer XL & 2.9B & 63.8 & 1.3& -1.49 & 87.8 &	81.5 &	89.5\\ 
       \hline
       Evolved Transformer Tiny & 19M & 1.31 & 39.7 & -2.45&69.6	&57.1 &	69.6\\ 
       Evolved Transformer Small & 79M & 4.23 & 23.7& -2.04 & 75.7 &	66.2 &	80.2\\ 
       Evolved Transformer Base & 218M & 10.2 & 8.9 &  -1.79&83.0 &	70.5 &	84.8\\
       Evolved Transformer Large & 1.0B & 49.3 & 2.1 &-1.62& 86.2 &	77.1 &	88.0\\
       Evolved Transformer XL & 2.2B & 71.3 & 0.8 &-1.55 &87.0	& 78.3 &	88.2\\ 
       \hline
       Universal Transformer Tiny& 11M & 1.77 & 38.1&-2.73 & 69.8	& 56.1 &	62.3\\
       Universal Transformer Small & 52M & 7.30&18.3 & -2.12 & 76.8	&64.2	& 75.4\\
       Universal Transformer Base & 127M & 20.3 & 8.4 &-1.91 & 80.0 &	67.9 &	80.1\\ 
       Universal Transformer Large & 283M & 27.6 & 1.6 &-1.67 & 84.0	& 73.4 & 	85.4\\
       \hline
       Switch Transformer Tiny & 174M & 3.25 & 29.7 & -2.01 & 78.2 &	63.8 &	80.7\\
       Switch Transformer Small & 460M & 4.63 & 22.3 & -1.85 & 80.3	& 68.0 &	82.9\\ 
       Switch Transformer Base & 2.0B & 12.7 &8.4 & -1.66 & 84.2 &	74.1 & 	86.5\\ 
       Switch Transformer Large & 3.9B & 23.0 & 4.1 & -1.56 & 84.6	& 75.8 &	87.9\\ 
       Switch Transformer XL & 29.6B & 43.3 & 0.8 &-1.62 &84.0 &	75.2&	87.5 \\
       \hline
       Performer Tiny & 16M & 1.14 & 42.0& -2.88 & 50.5 &	48.8 & 	15.0\\
       Performer Small & 61M & 3.50 & 39.0 & -2.44 & 57.8	& 51.1 &	31.1\\ 
       Performer Base & 224M & 10.8 & 11.7 & -2.23 & 61.4 & 	53.4 &	37.8\\ 
       Performer Large & 739M & 32.8 & 4.4 &-2.16 & 62.4 &	52.4 &	30.8\\ 
       \hline
       Funnel Transformer Tiny & 16M & 1.10 & 39.9& -2.58 & 63.4 &	49.4	& 54.6\\
       Funnel Transformer Small & 61M & 2.96 &32.7 &-2.11 & 70.0 & 	58.5 &	75.1\\ 
       Funnel Transformer Base & 223M & 8.10 & 11.9& -1.83 & 76.3 &	62.9 & 	81.6\\ 
       Funnel Transformer Large & 739M & 22.6 & 5.0 & -1.69 & 79.8 &	67.1 & 	83.8\\
       Funnel Transformer XL & 2.9B & 40.3 & 1.89 &-1.61 & 79.8 &	68.0 &	83.7\\ 
       \hline
       ALBERT Small & 15M & 3.57 & 42.0 &-2.36 & 73.7 &	62.0 &	77.1\\ 
       ALBERT Base & 21M & 9.40 & 16.4 &-2.28 & 69.0 &	57.2 &	64.3\\ 
       ALBERT Large & 34M & 31.6 & 5.1&-2.20 & 62.9	& 54.1 &	27.3\\ 
       \hline
       MoS-Transformer Tiny & 27M & 1.29 &39.7 & -2.37 & 70.6 & 57.9 & 	74.1\\
        MoS-Transformer Small & 81M & 3.70 & 26.3 &-1.98 & 79.7 &	67.1 &	83.1\\
        MoS-Transformer Base & 257M & 11.4 & 8.6&-1.70 & 84.5 &	73.9 &	86.8\\
        MoS-Transformer Large & 800M & 35.0 & 3.4&-1.56 & 86.5 &	79.7 &	89.1\\ 
        MoS-Transformer XL & 2.9B &  112 & 1.2 &-1.45 & 88.2 &	81.4 &	90.0\\
        \hline
        GLU-Transformer Tiny & 26M & 1.29 & 31.7 & -2.35 & 70.5 &	57.0 &	74.2\\
        GLU-Transformer Small & 77M & 3.70 & 26.4 & -1.97 & 79.1	& 67.4 &	83.0\\ 
        GLU-Transformer Base & 248M & 11.4 & 8.6&  -1.71 & 84.6 &	74.5 &	87.2\\
        GLU-Transformer Large & 748M & 35.0 & 3.4 & -1.56&84.2 &	74.3 & 	86.2\\ 
        GLU-Transformer XL & 2.85B &  61.3   & 1.0 &-1.49 & 87.6 &	82.9 &	89.4\\ 
        \hline
        LConv Tiny & 17M & 1.20 & 31.2 & -2.50 & 51.1 &	51.3 &	49.5\\
        LConv Small & 67M & 3.80 & 12.8&-2.10 & 71.8 &	59.9 &	64.7\\ 
        LConv Base &  210M & 10.6 & 12.8 &-1.95 & 73.8 &	63.6 &	70.3 \\ 
        LConv Large &  741M & 41.0 &3.0 & -1.76 & 76.8 &	65.6 &	76.3\\ 
        LConv XL & 2.3B & 77.0 & 1.0 & -1.75 & 73.3 &	64.1 &	72.9\\ 
        \hline
        DConv Tiny &  22M & 1.39 & 27.3&-2.46 & 51.1 &	48.9 &	30.2 \\ 
        DConv Small & 96M & 4.97 & 19.8 &-2.08 & 68.6	& 57.4	& 64.3\\ 
        DConv Base & 324M & 15.3 & 7.6 &-1.90 & 72.9	& 60.1 &	63.7\\ 
        DConv Large & 1.2B & 78.0 & 1.1 &-1.82 & 70.8	&58.5 &	58.2\\ 
        \hline
        MLP-Mixer Small & 67M & 3.83 & 22.3 & -2.15 & 65.4	& 55.1 &	58.7 \\ 
        MLP-Mixer Base & 233M & 12.4 & 10.7& -1.90 & 64.4 &	58.1 &	60.5\\ 
        MLP-Mixer Large & 739M & 38.3 & 3.9 & -1.73 & 52.2 &	47.8 &	60.9\\ 
        MLP-Mixer XL & 2.86B & 48.3 & 1.2 &-1.61 & 57.3 &	58.9 &	65.7\\ 
        \hline
     \end{tabular}}
  
     \label{tab:all_results}
 \end{table*}
\subsection{Scaling Law for Each Model}
Table \ref{tab:slope} presents the slope of the fitted linear line $\alpha$ for each model across multiple scenarios. We derive $\alpha$ by plotting $F$ (FLOPs), $U$ (upstream perplexity), $D$ (downstream accuracy), $P$ (number of parameters). In general, most values of $\alpha$ depict how well a model scales. For example $\alpha_{F,U}$ is plotting FLOPs against Upstream performance. The only exception is $\alpha_{U,D}$ which is a measure of upstream vs downstream performance. A high $\alpha_{U,D}$ value means that the transfer to the downstream tasks is better as a model scales. Overall, the $\alpha$ value is a metric that represents how well a model performs relatively across all scales 

\begin{table}[t]
    \centering
    \small
    \caption{Slope of a fitted linear line for each model, when we compare FLOPs vs. upstream performance ($F,U$), FLOPs vs. downstream performance ($F,D$), parameter size vs. upstream performance ($F,U$), parameter size vs. downstream performance ($P,D$), and finally upstream performance  vs. downstream performance ($U,D$).}
    \begin{tabular}{l|ccccc}
    \hline
       Model  &  $\alpha_{F,U}$ & $\alpha_{F,D}$ & $\alpha_{P,U}$ & $\alpha_{P,D}$ & $\alpha_{U, D}$ \\
       \hline
       Transformer     & \textbf{0.54}& \textbf{0.28} & \textbf{0.47} & \textbf{0.24} & 0.49 \\
       GLU-Trans. & 0.49 & 0.24 & 0.42 & 0.22 & 0.46 \\
       LConv           & 0.32 & 0.13 & 0.29 & 0.11 & 0.48 \\
       Funnel        & 0.47 & 0.22 & 0.38 & 0.18 & 0.46 \\ 
       Switch         & 0.23 & 0.14 & 0.13 & 0.08 & \textbf{0.58} \\ 
       Universal            & 0.50 & 0.20 & 0.56 & 0.22 & 0.35 \\ 
       ALBERT          & 0.08 & -0.12 & 0.13 & -0.21 & -1.67 \\ 
       Evolved      & 0.44 & 0.22 & 0.42 & 0.21 & 0.47 \\
       Performer       & 0.25 & 0.05 & 0.24 & 0.05 &  0.24 \\ 
       MoS-Trans. & 0.43 & 0.21 & 0.43 & 0.20 & 0.47 \\
       MLP-Mixer       & 0.32 & -0.03 & 0.26 & 0.65 & -0.02 \\ 
         \hline
    \end{tabular}
    
    \label{tab:slope}
\end{table}

\paragraph{Analysis of Slope for each Model} In general, we find that the vanilla Transformer has the highest values of $\alpha$. Models such as Evolved Transformer, GLU-Transformer, MoS-Transformer and Funnel Transformer tend to have similar scaling properties to the vanilla Transformer. The GLU-Transformer has similar and slightly worse scaling properties to the vanilla Transformer, even if it was observed to do better in absolute sense on some compute-regions. On the other hand, we also observe that there are models which are difficult to scale such as LConv, UT, MLP-Mixer and Performer. This is even more evident on downstream task. We also note that ALBERT scales (trends) negatively\footnote{This version of ALBERT shares parameters across encoder and decoder which may partially explain why we had a hard time scaling up.} (gets worse) as we scale the model up.  On the other hand, the metric $\alpha_{U,D}$ measures how the downstream performance scales with upstream performance. Overall, the Switch Transformer does the best on this metric where downstream performance scales well with upstream performance. Generally, models that make less changes to the main Transformer architecture (GLU-Transformer, MoS-Transformer) tend to retain similar scaling behaviours and changing the inductive bias also significantly alters the scaling property of the model.



\subsection{Do Scaling Protocols influence model architectures in the same way?}
\begin{figure}[t]
     \centering
     \begin{subfigure}{0.24\textwidth}
         \centering
         \includegraphics[width=\textwidth]{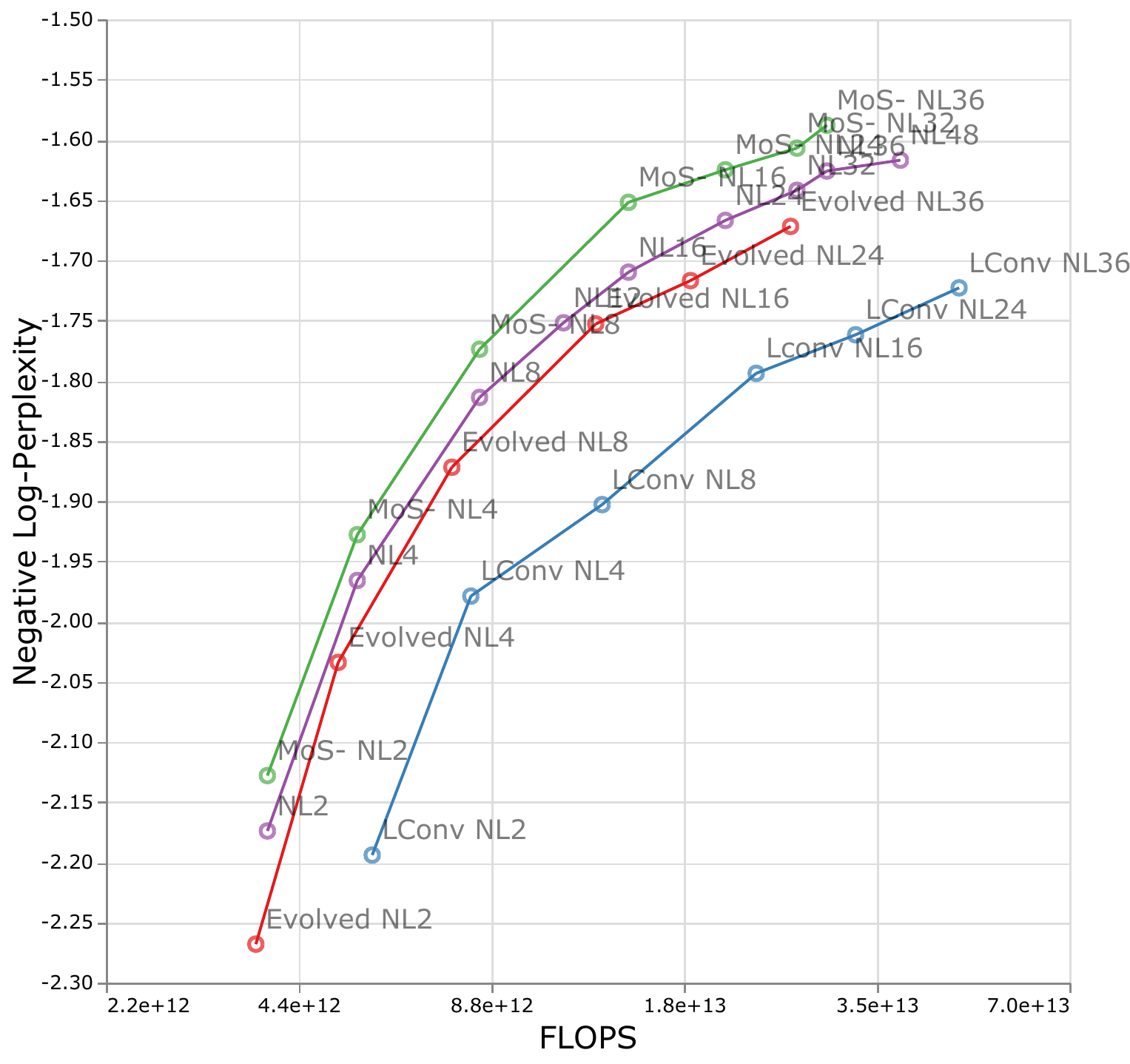}
         \caption{Upstream Neg. Log-PPL.}
         \label{fig:scaling_depth_us}
     \end{subfigure}%
     \begin{subfigure}{0.24\textwidth}
         \centering
         \includegraphics[width=\textwidth]{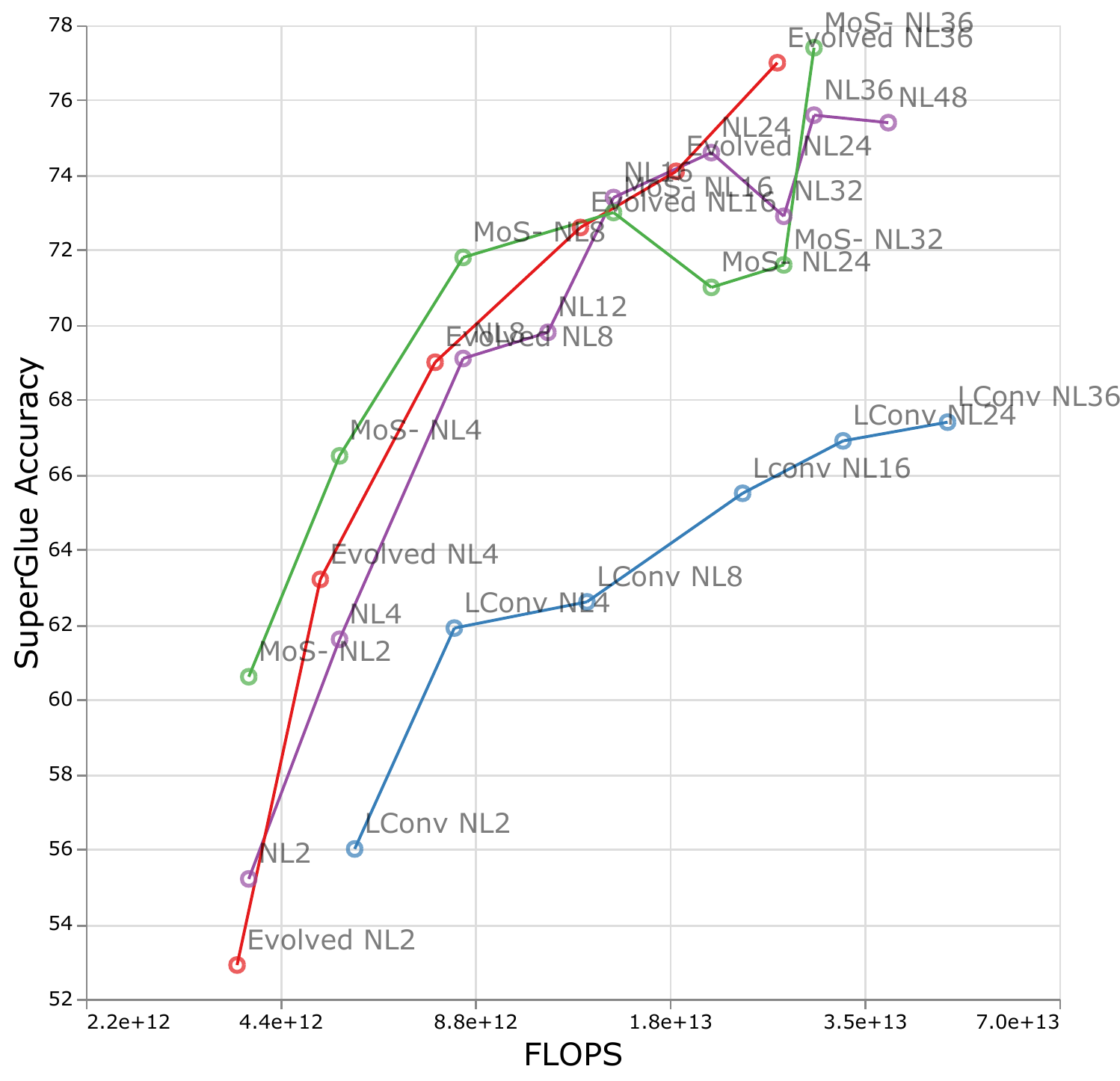}
         \caption{Downstream Accuracy.}
         \label{fig:scaling_depth_ds}
     \end{subfigure}
    \caption{Scaling depth}
    \label{fig:scaling_depth}
\end{figure}

\begin{figure}[t]
     \centering
     \begin{subfigure}{0.24\textwidth}
         \centering
         \includegraphics[width=1.0\textwidth]{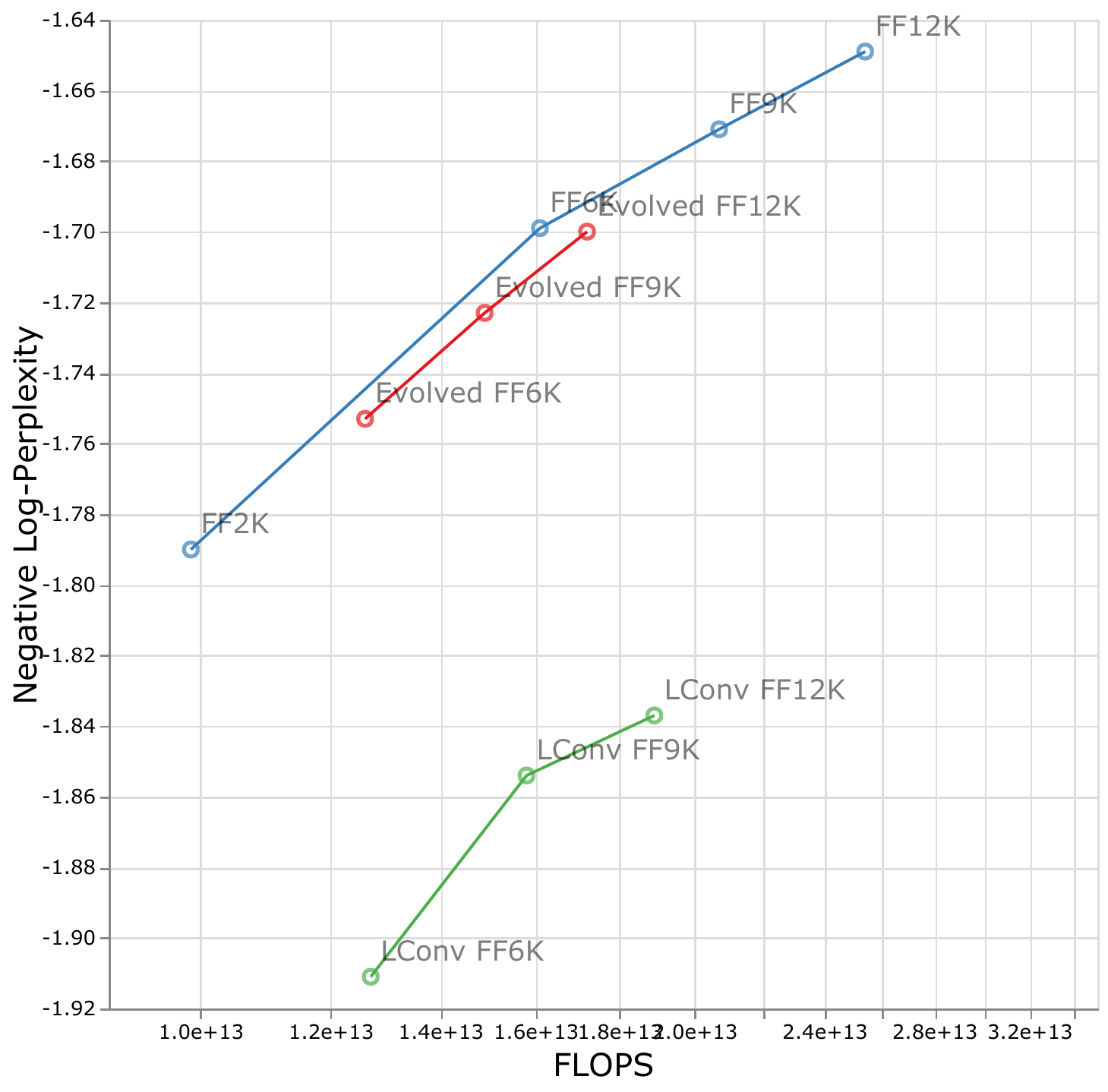}
         \caption{Upstream Neg. Log-PPL.}
         \label{fig:scaling_width_us}
     \end{subfigure}%
     \begin{subfigure}{0.24\textwidth}
         \centering
         \includegraphics[width=1.0\textwidth]{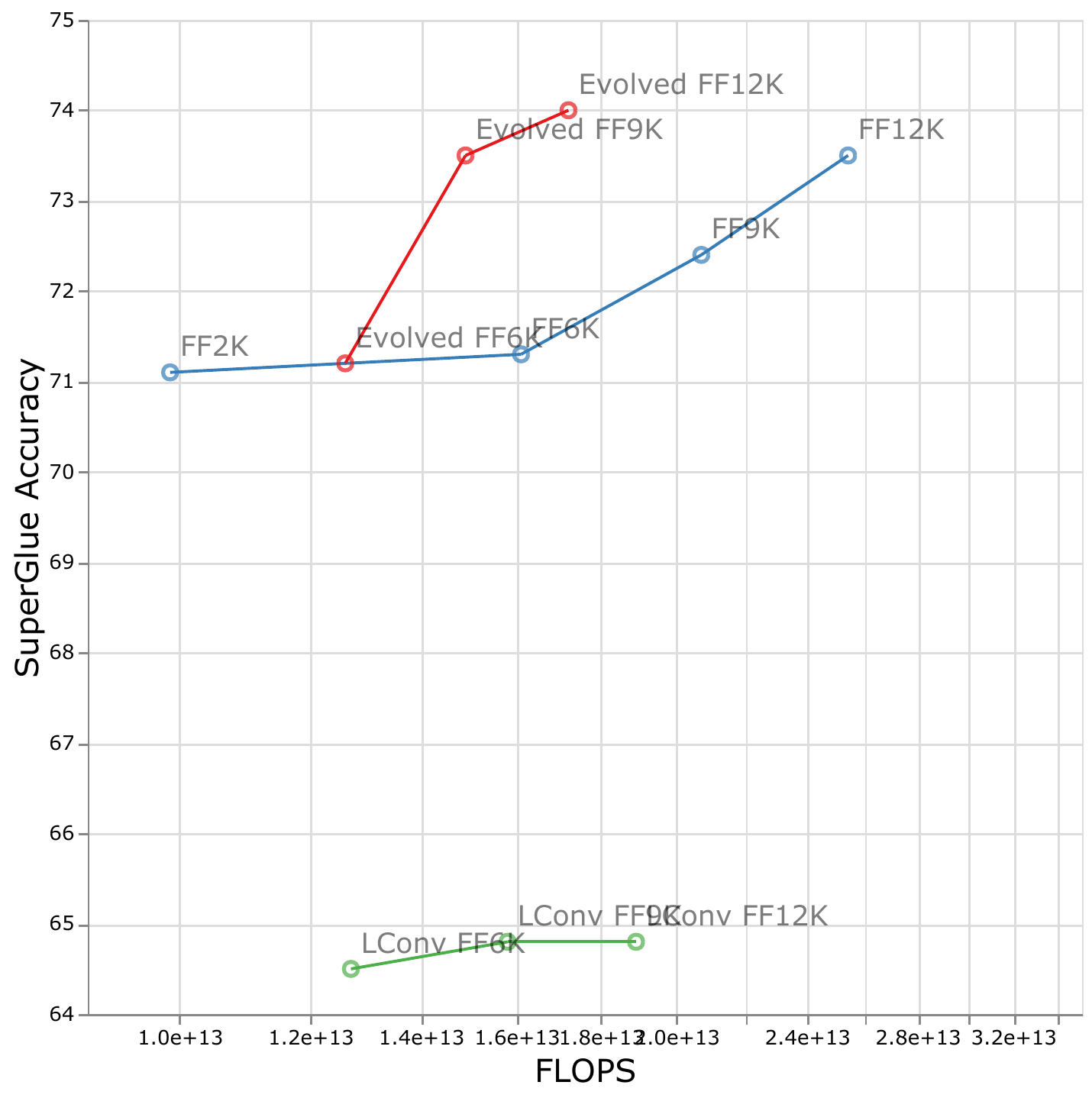}
         \caption{Downstream Accuracy.}
         \label{fig:scaling_width_ds}
     \end{subfigure}
    \caption{Scaling width of FFN}
    \label{fig:scaling_width}
\end{figure}

We are interested in how different scaling protocols influence the model architectures. Figure \ref{fig:scaling_depth} shows the effect of scaling depth of four model architectures (MoS-Transformer, Transformer, Evolved Transformer and LConv). Figure \ref{fig:scaling_width} shows the effect of scaling width on the same four architectures. Firstly, on upstream (negative log perplexity) curves, we note that while different architectures have a distinct difference in absolute performance, the scaling trend remains quite similar. On downstream, depth scaling (Figure \ref{fig:scaling_depth}) seems to act equally on most architectures with the exception of LConv. Meanwhile, for width scaling, it seems that Evolved Transformers scale slightly better when applying width-scaling. It is also interesting to note that depth-scaling has a much more substantial impact on downstream scaling as opposed to width-scaling.

\subsection{Epilogue and Conclusion}
In this paper, we conducted extensive experiments, pretraining and finetuning of up to 100 models ranging from 10 well-established Transformer and non-Transformer architectures. We showed that different model architectures can have different scaling behaviours and models performing well in one compute region (or model size) may not do identically well in another compute region.

We also showed that model architectures may do well on upstream perplexity but fail to transfer to downstream tasks. Hence, practitioners should be cautious about developing architectures that not only scale well with respect to the upstream perplexity, but also based on downstream performance. 
While we certainly do not expect researchers to always report model performance across all scales (especially large-scale), we believe that it is good to keep in mind that architectures can perform quite differently at different compute regions. 
Hence, this might be a good dimension to consider when designing new inductive biases. As such, performing evaluation at a certain compute region may be insufficient to capture the full picture. It is also good to consider if different inductive biases will result in different extends of emergent capabilities \citep{wei2022emergent, abnar2020transferring}.

We also showed that different model architectures may react differently to different scaling protocols, which further expands on the narrative that comparing and benchmarking these models can be very challenging~\citep{dehghani2021benchmark}. When it comes to scaling large models, this paper shows that novel inductive biases can be indeed quite risky which might explain why most state-of-the-art large language models \cite{rae2021scaling,chowdhery2022palm,tay2022unifying} are based on relatively vanilla architectures.  Our advice is to be cautious when staking an expensive run on a Transformer architecture that drastically modifies the attention mechanism (e.g., Mixers and Performers are generally high risk options as seen in our experiment results).
Finally, we acknowledge that not every practitioner or researcher would require models that are able to scale to billion of parameters. In that case, inductive biases that are tailored to small or low compute will be sufficient.

\bibliography{custom}
\bibliographystyle{acl_natbib}

\newpage

\section{Appendix}

\subsection{Scaling Details for Individual Models}
For most models, it was reasonable to follow the uniform scaling method in the main T5 sizes. At each size, the hyperparameters are as follows: 

  \begin{table}[H]
    \centering
    \small
    \begin{tabular}{cccccccc}
    \toprule
      Model & $N_L$ & $d_{ff}$ & $d_{model}$ & $d_{kv}$ & $N_H$ &  \#Params  \\
      \hline
      Tiny &4/4 & 1024 & 256 & 32 & 4  & 16M \\ 
      Small & 6/6 & 2048 & 512 & 32 & 8  &  60M  \\
     Base & 12/12 & 3072 & 768 &64 & 12 &  220M \\
     Large & 24/24 & 4096 & 1024 & 64 & 16  & 738M\\ 
     XL & 24/24 & 16384 & 1024 & 128 & 32 & 3B \\ 
         \hline
    \end{tabular}
    \caption{Table of model configurations. $N_L$ is the number of layers, $d_{ff}$ is the size of the MLP, $d_{model}$ is the hidden size of the model. $d_{kv}$ is the size of each key-value vector. $N_H$ is the number of heads. }
    \label{tab:my_label-1}
\end{table}

\paragraph{Scaling for Switch Transformer} For Switch Transformers, we use the following scaling:
\begin{table}[H]
    \centering
    \small
    \begin{tabular}{lcccccccc}
    \hline
      Model  & $N_L$ & $d_{ff}$ & $d_{model}$ & $d_{kv}$ & $N_H$ & $N_{E}$ &\#Params  \\
      \hline
      Tiny  & 4 & 1024 & 512 & 64 & 12 & 32 & 173M \\ 
      Small & 6 & 2048 & 512 & 64 & 12 & 32 & 460M\\ 
     Base & 12 & 3072 & 768 & 64 & 12 & 32 & 2B\\ 
        Large & 24 & 3072 & 768 & 64 & 12 & 32 & 8B\\
       XL & 48 & 3072 & 768 & 64 & 12 & 128 & 30B \\ 
      \hline
    \end{tabular}
    \caption{Scaling for Switch Transformer. $N_E$ is the number of experts.}
    \label{tab:my_label-2}
\end{table}
\paragraph{Scaling for Universal Transformer} Scaling UTs are generally difficult as described in the main text. There were two main considerations for scaling UTs. Initially we tried scaling the number of recurrent operations. However, we found that even with an increase of FLOPS, this does not lead to improved performance. Overall, the UT model might be pretty slow and therefore a model with the same hparams as vanilla XL might be infeasible to run. Hence, we explored increasing the width of the MLPs to $32K$ to see if UTs would scale in this manner.
 \begin{table}[H]
    \centering
    \small
    \begin{tabular}{lccccccc}
    \hline
      Model & $N_R$ & $d_{ff}$ & $d_{model}$ & $d_{kv}$ & $N_H$ &  \#Params  \\
      \hline
      UT Tiny & 3/3 & 1024 & 128 & 32 & 8  & 11M \\ 
      UT Small & 3/3 & 2048 & 512 & 32 & 8  &  52M  \\
    UT  Base & 3/3 & 3072 & 768 &64 & 12 &  127M \\
   UT   Large & 3/3 & 32768 & 1024 & 64 & 16  & 283M\\ 
         \hline
    \end{tabular}
    \caption{Table of model configurations. $N_R$ is the number of recurrent operations, $d_{ff}$ is the size of the MLP, $d_{model}$ is the hidden size of the model. $d_{kv}$ is the size of each key-value vector. $N_H$ is the number of heads. }
    \label{tab:my_labe-3}
\end{table}

\newpage
\begin{figure*}[t]
     \centering
    \begin{subfigure}[b]{\textwidth}
         \centering
         \includegraphics[width=0.7\textwidth]{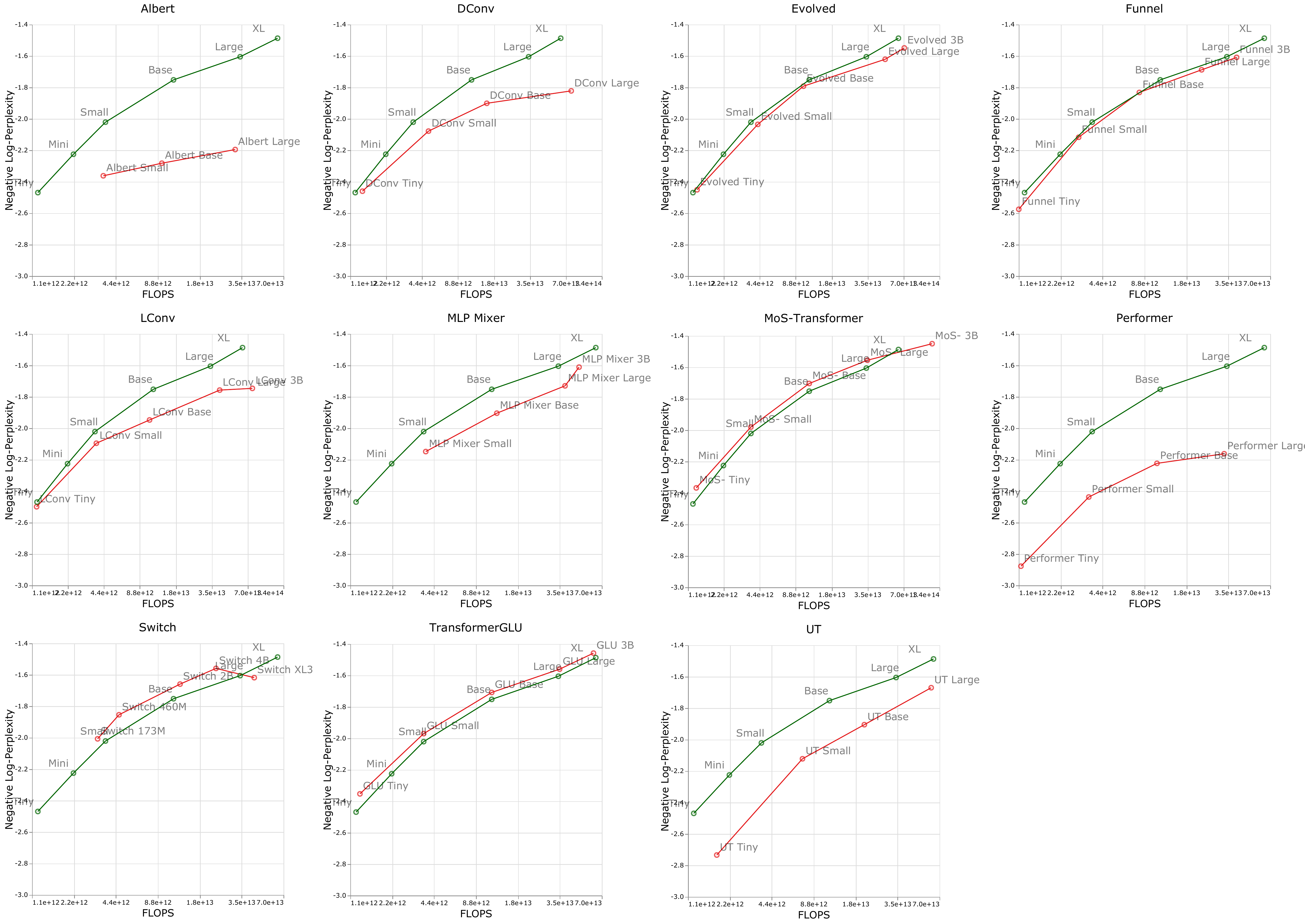}
         \caption{FLOPs}
         \label{fig:all_models_ppl_flops}
     \end{subfigure}
     \begin{subfigure}[b]{\textwidth}
         \centering
         \includegraphics[width=0.7\textwidth]{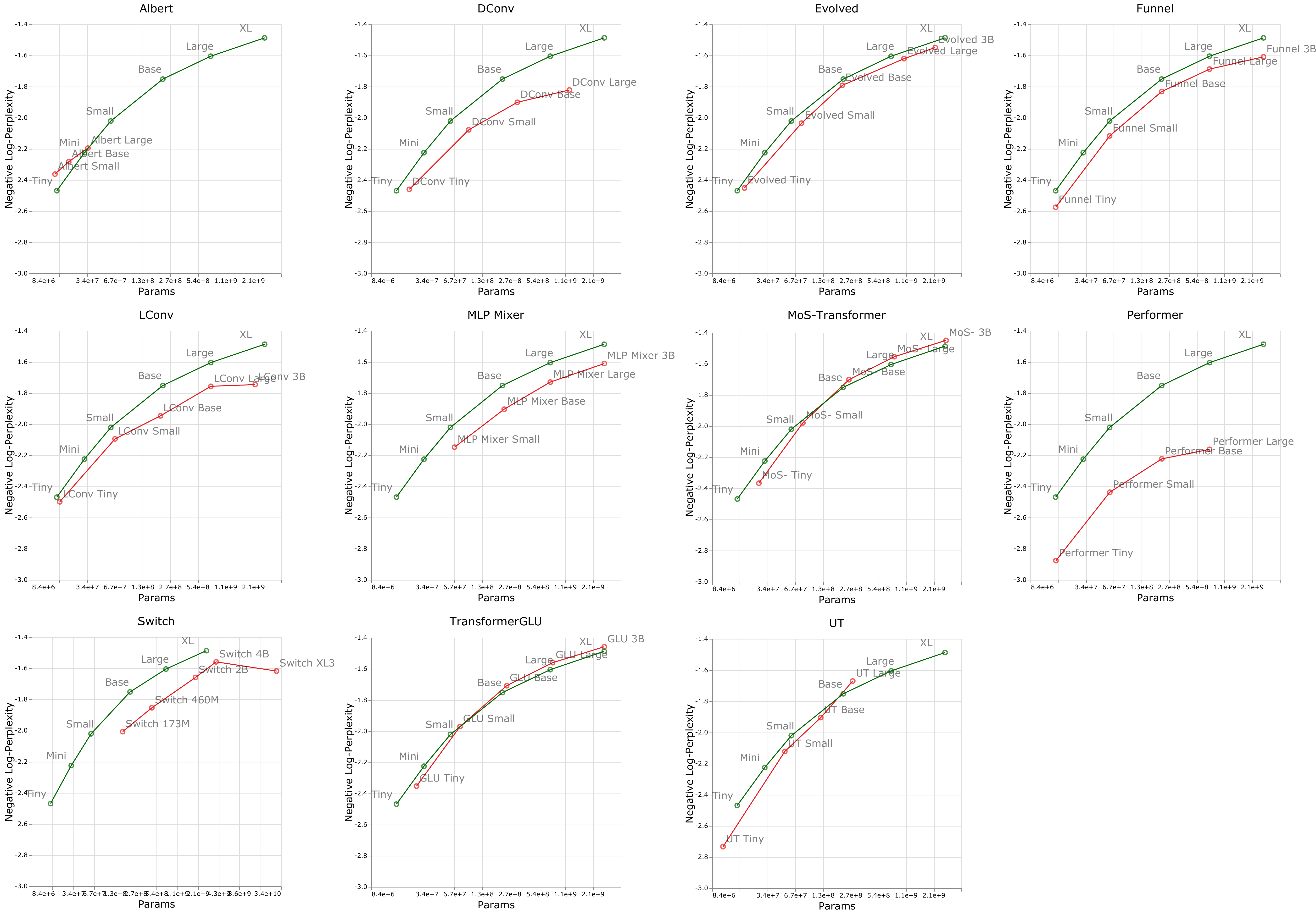}
         \caption{Number of Parameters}
         \label{fig:all_models_ppl_params}
     \end{subfigure}
     \begin{subfigure}[b]{\textwidth}
         \centering
         \includegraphics[width=0.7\textwidth]{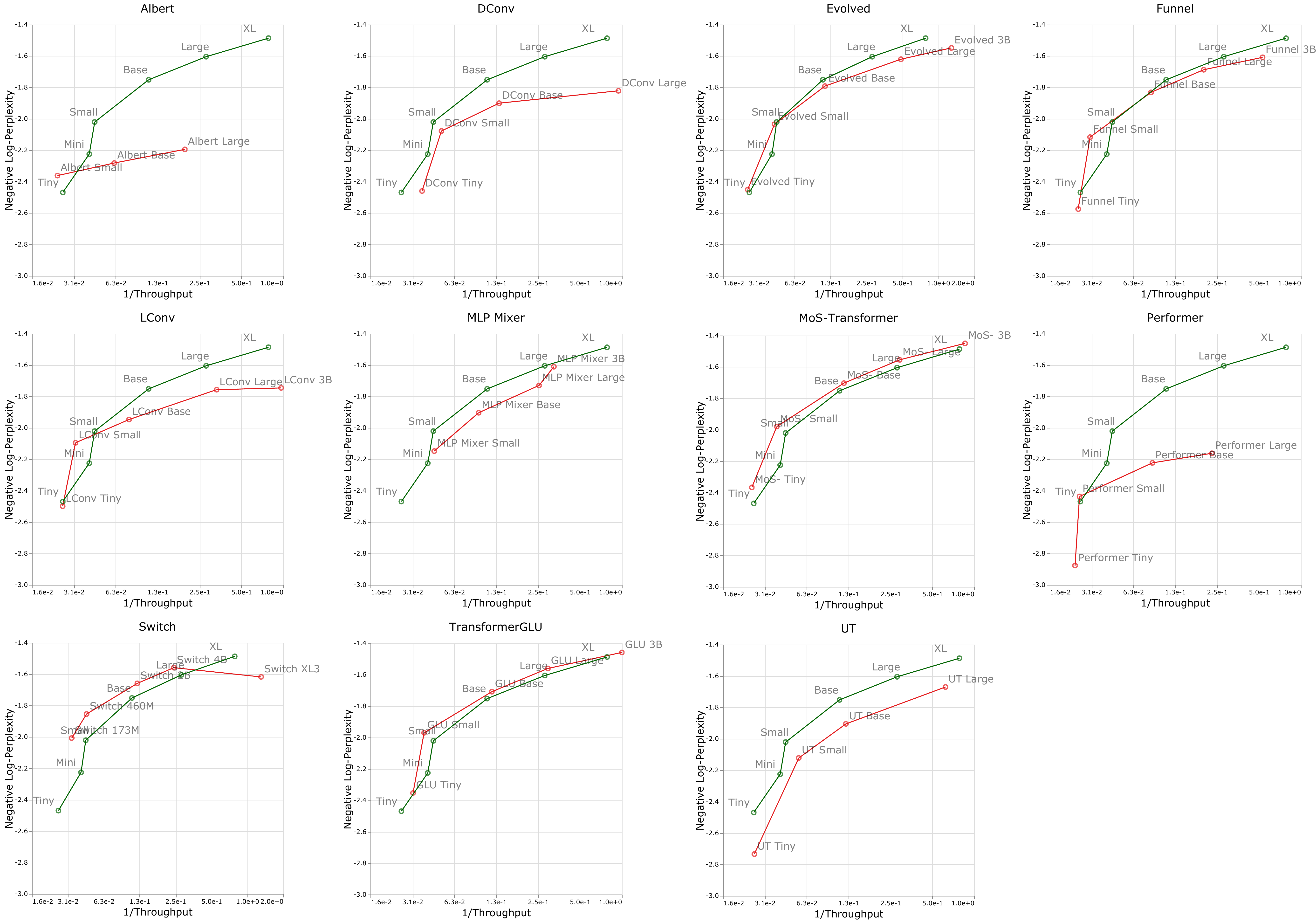}
         \caption{Throughput}
         \label{fig:all_models_ppl_throughput}
     \end{subfigure}
    \caption{Quality-cost trade of for the upstream Negative Log-Perplexity of vanilla Transformer compared to other models, with respect to FLOPs, number of parameters, and throughput.}
    \label{fig:all_models_ppl}
\end{figure*}

\begin{figure*}[t]
     \centering
    \begin{subfigure}[b]{\textwidth}
         \centering
         \includegraphics[width=0.7\textwidth]{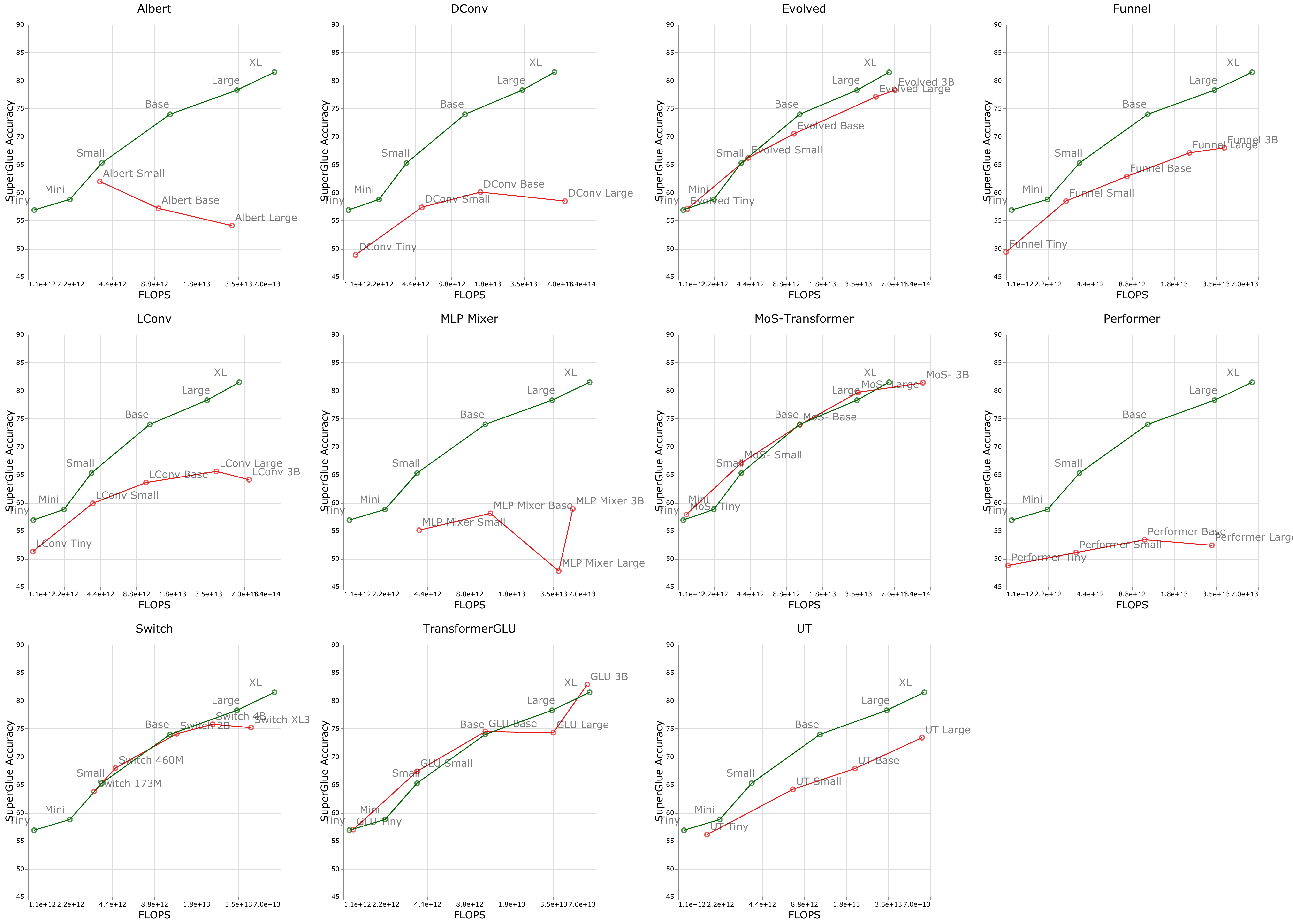}
         \caption{FLOPs}
         \label{fig:all_models_sga_flops}
     \end{subfigure}
     \begin{subfigure}[b]{\textwidth}
         \centering
         \includegraphics[width=0.7\textwidth]{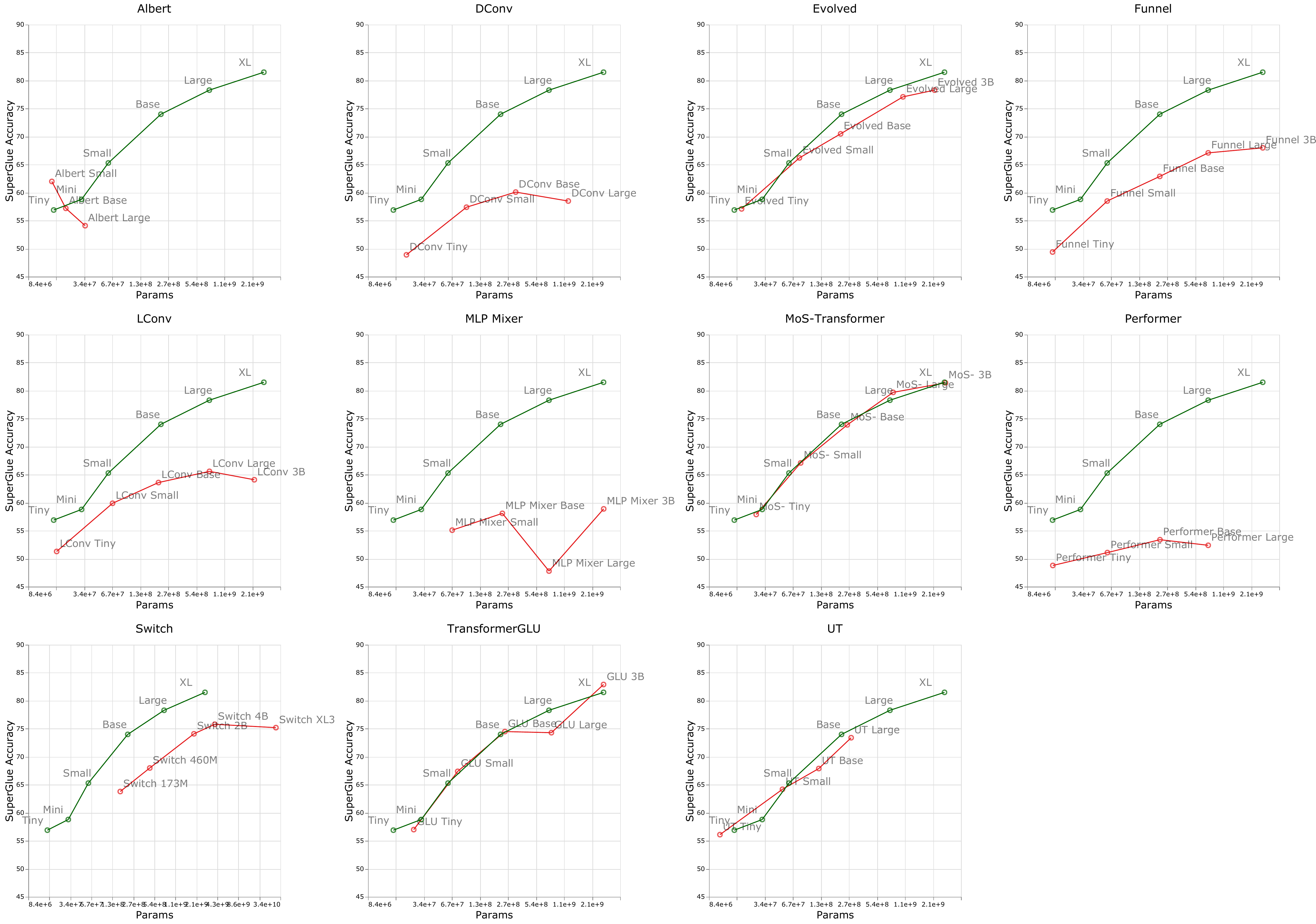}
         \caption{Number of Parameters}
         \label{fig:all_models_sga_params}
     \end{subfigure}
     \begin{subfigure}[b]{\textwidth}
         \centering
         \includegraphics[width=0.7\textwidth]{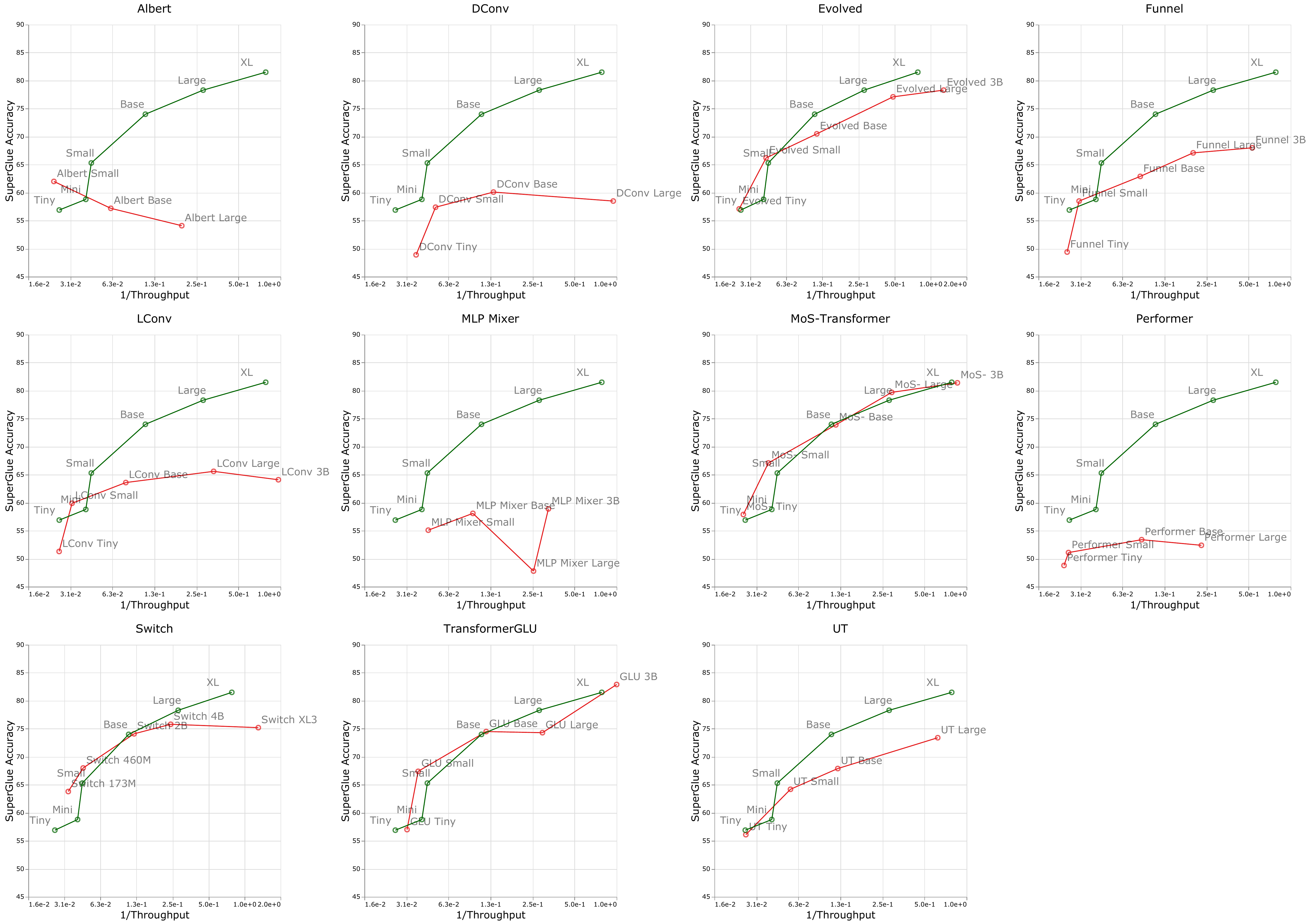}
         \caption{Throughput}
         \label{fig:all_models_sga_throughput}
     \end{subfigure}
    \caption{Quality-cost trade of for the downstream SuperGlue Accuracy of vanilla Transformer compared to other models, with respect to FLOPs, number of parameters, and throughput.}
    \label{fig:all_models_sga}
\end{figure*}

\begin{figure*}[t]
     \centering
    \begin{subfigure}[b]{\textwidth}
         \centering
         \includegraphics[width=0.6\textwidth]{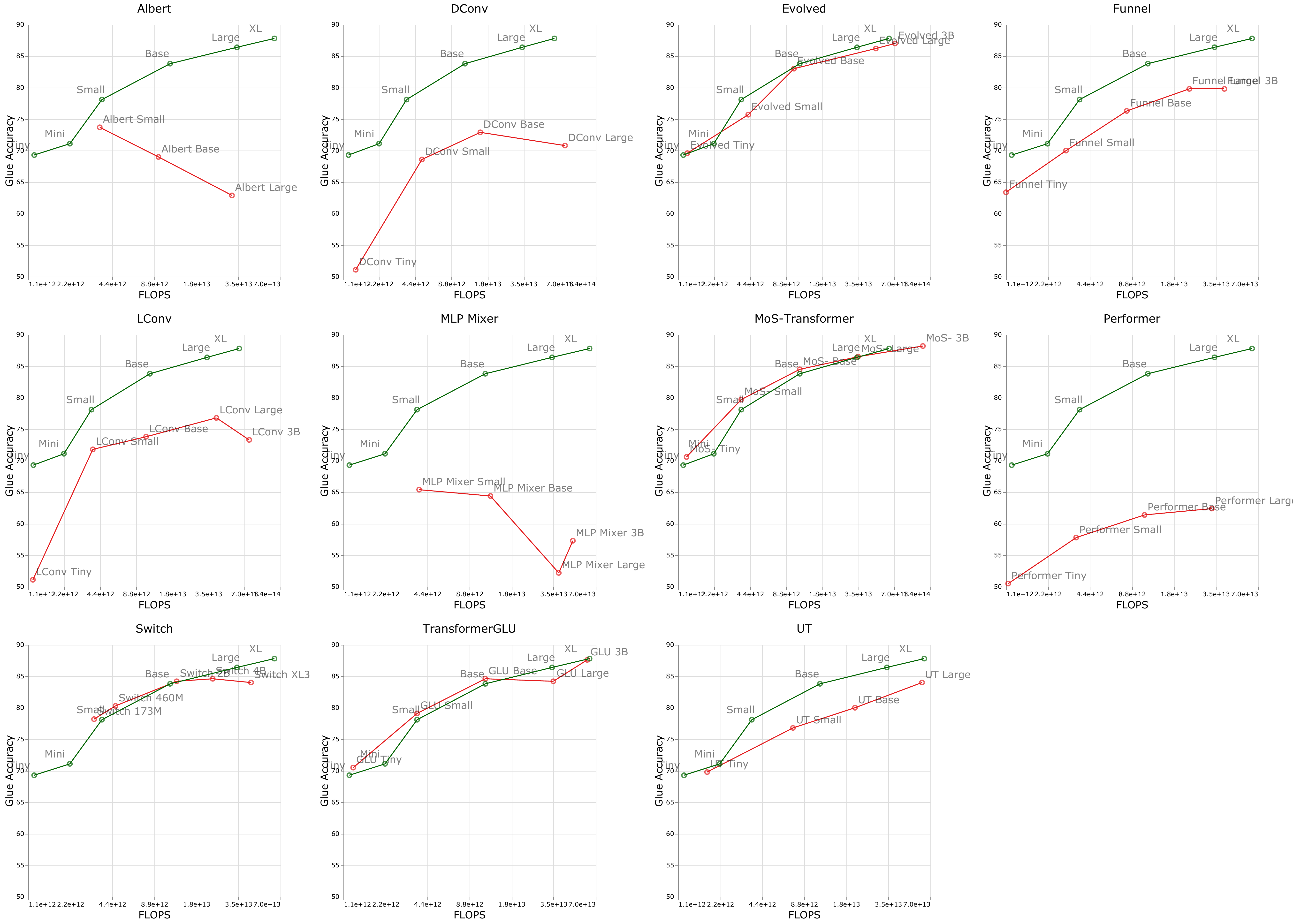}
         \caption{FLOPs}
         \label{fig:all_models_ga_flops}
     \end{subfigure}
     \begin{subfigure}[b]{\textwidth}
         \centering
         \includegraphics[width=0.6\textwidth]{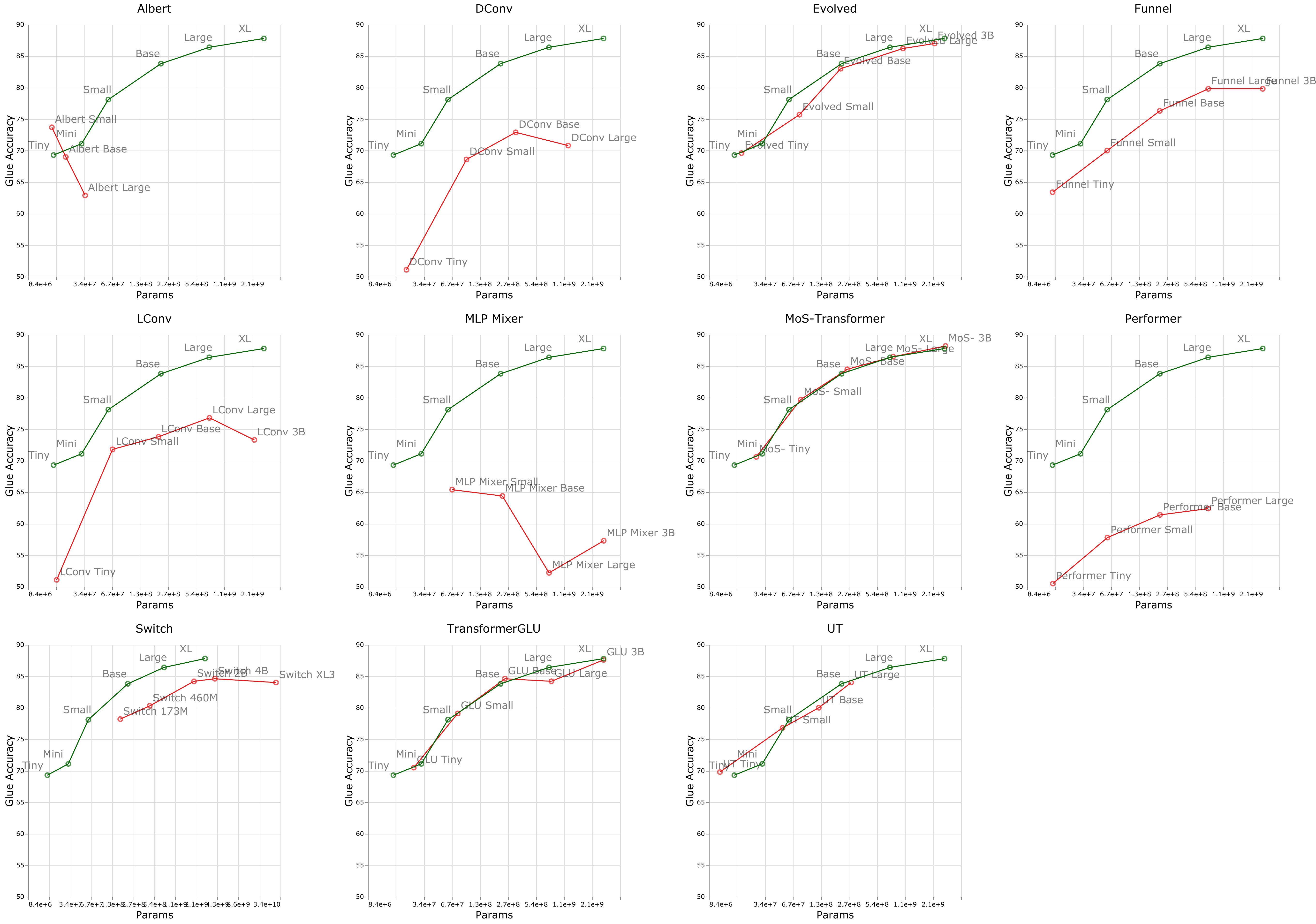}
         \caption{Number of Parameters}
         \label{fig:all_models_ga_params}
     \end{subfigure}
     \begin{subfigure}[b]{\textwidth}
         \centering
         \includegraphics[width=0.6\textwidth]{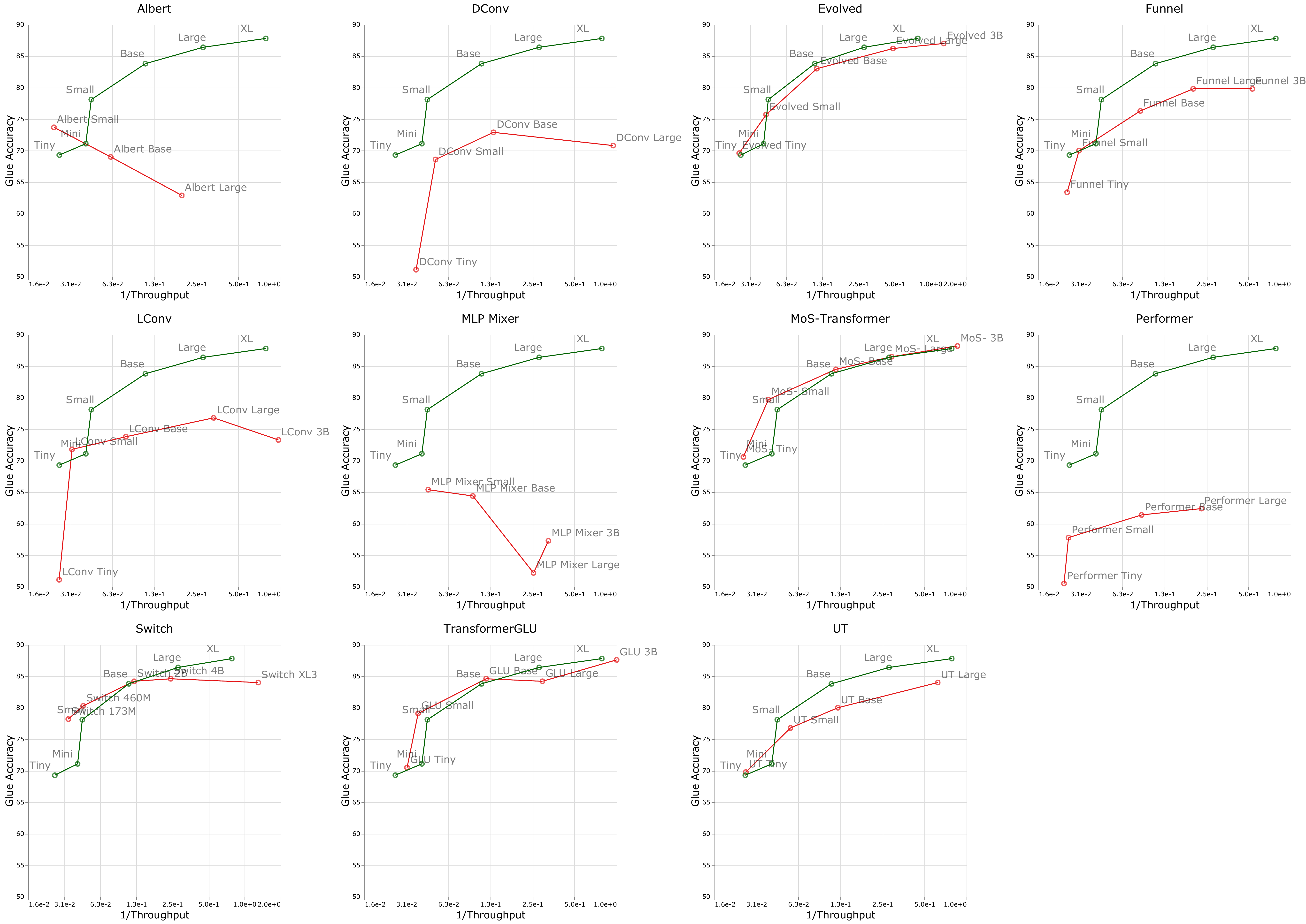}
         \caption{Throughput}
         \label{fig:all_models_ga_throughput}
     \end{subfigure}
    \caption{Quality-cost trade of for the downstream Glue Accuracy of vanilla Transformer compared to other models, with respect to FLOPs, number of parameters, and throughput.}
    \label{fig:all_models_ga}
\end{figure*}

\begin{figure*}[t]
\vspace{-50pt}
     \centering
    \begin{subfigure}[b]{\textwidth}
         \centering
         \includegraphics[width=0.7\textwidth]{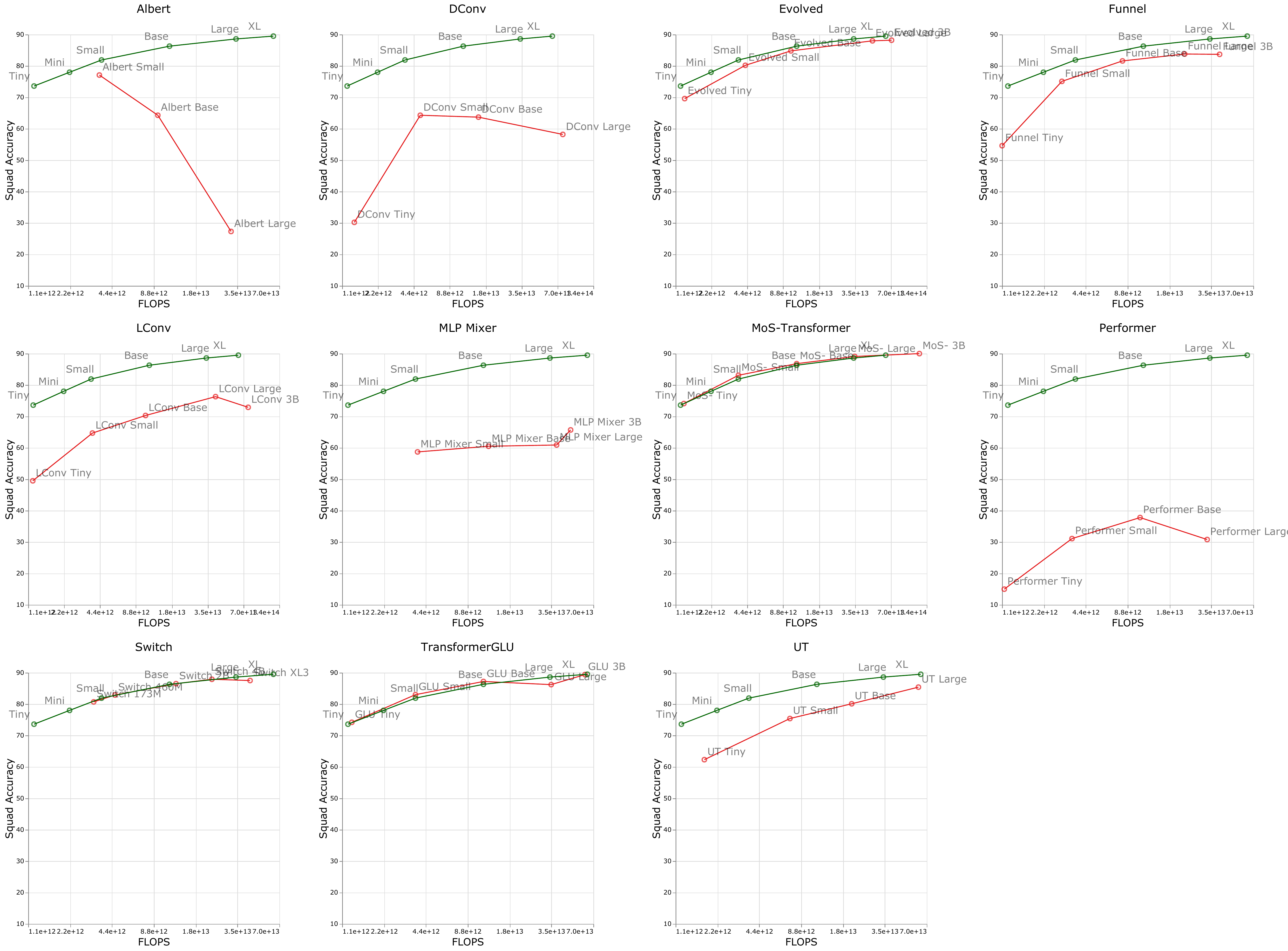}
         \caption{FLOPs}
         \label{fig:all_models_sa_flops}
     \end{subfigure}
     \begin{subfigure}[b]{\textwidth}
         \centering
         \includegraphics[width=0.7\textwidth]{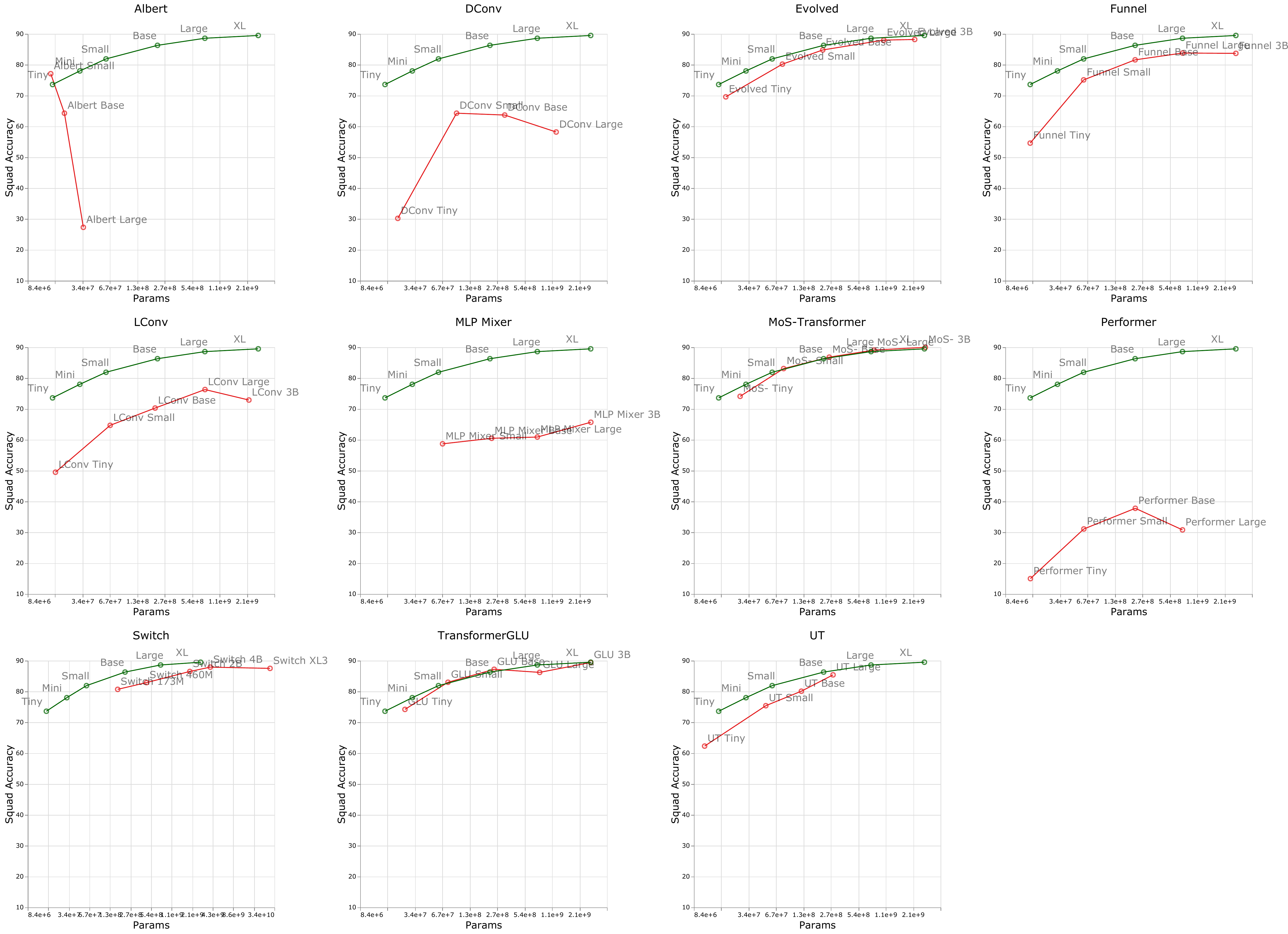}
         \caption{Number of Parameters}
         \label{fig:all_models_sa_params}
     \end{subfigure}
     \begin{subfigure}[b]{\textwidth}
         \centering
         \includegraphics[width=0.7\textwidth]{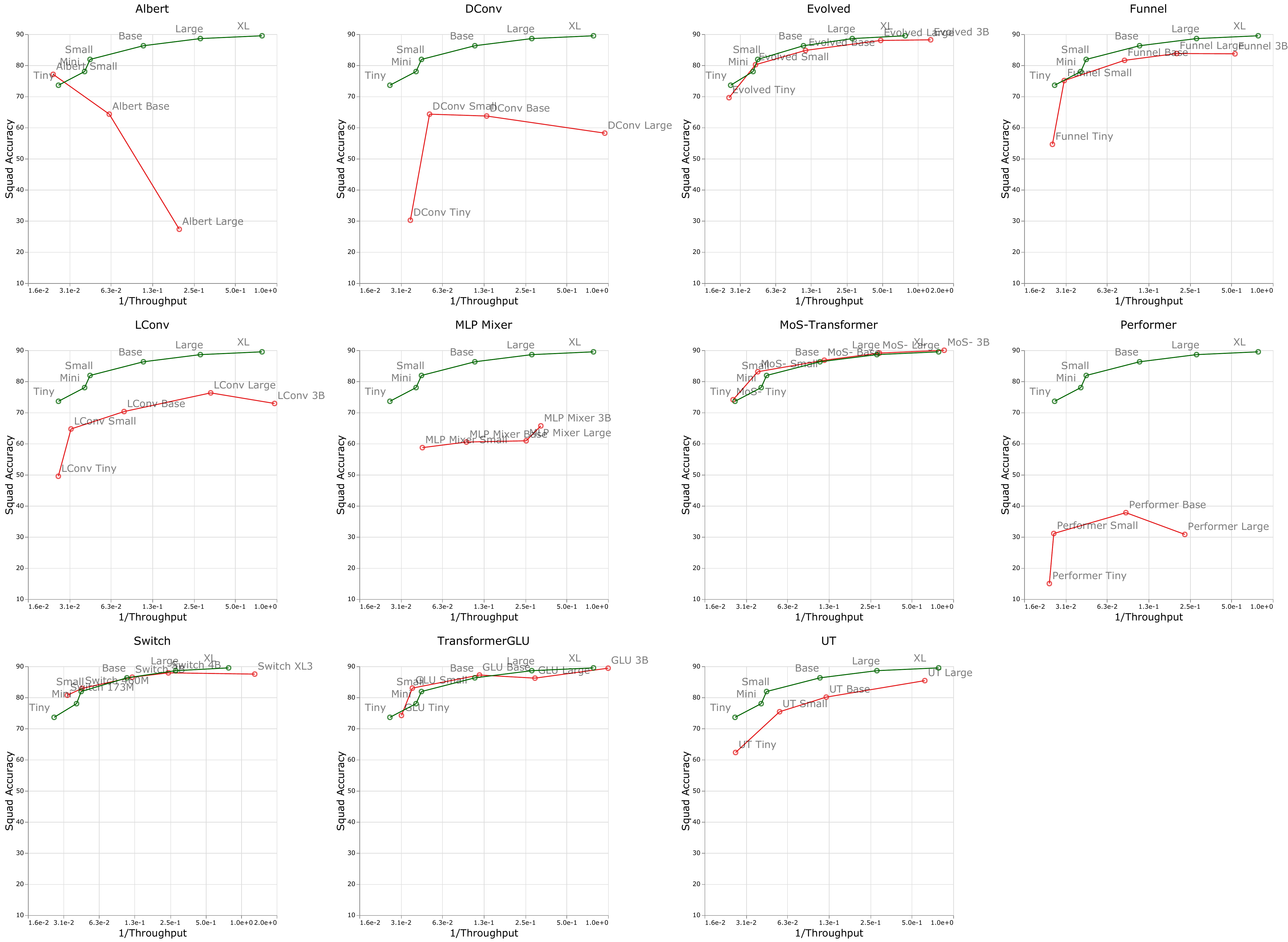}
         \caption{Throughput}
         \label{fig:all_models_sa_throughput}
     \end{subfigure}
    \caption{Quality-cost trade of for the downstream Squad Accuracy of vanilla Transformer compared to other models, with respect to FLOPs, number of parameters, and throughput.}
    \label{fig:all_models_sa}
\end{figure*}
\end{document}